%% file: articleneurips.tex
\documentclass{article}

\pdfoutput=1 
\PassOptionsToPackage{numbers,sort,compress}{natbib}
\usepackage[preprint]{neurips_2022}

\usepackage[utf8]{inputenc} % allow utf-8 input
\usepackage[T1]{fontenc}    % use 8-bit T1 fonts
\usepackage{hyperref}       % hyperlinks
\usepackage{url}            % simple URL typesetting
\usepackage{booktabs}       % professional-quality tables
\usepackage{amsfonts}       % blackboard math symbols
\usepackage{nicefrac}       % compact symbols for 1/2, etc.
\usepackage{microtype}      % microtypography
\usepackage{xcolor}         % colors
\usepackage{amsmath}
\usepackage{xcolor}
\usepackage{tikz}
\usetikzlibrary{backgrounds,calc,shadings,shapes.arrows,arrows,shapes.symbols,shadows,positioning,decorations.markings,backgrounds,arrows.meta,positioning}
\usepackage{optidef}
\usepackage{multirow}
\usepackage{graphicx}

\usepackage{pgfplots} % added during rebuttal

\usepackage[ruled,vlined]{algorithm2e} %linesnumbered

\usepackage{enumitem}
\setitemize{itemsep=2pt,topsep=0pt,parsep=0pt,partopsep=0pt,leftmargin=*}
\def\aquad{\hskip0.4em\relax}

\title{Self-Supervised Learning with an Information Maximization Criterion}

\author{Serdar Ozsoy\textsuperscript{1,2} \aquad Shadi Hamdan\textsuperscript{1,3} \aquad Sercan Ö. Arik\textsuperscript{4} \aquad Deniz Yuret\textsuperscript{1,3} \quad Alper T. Erdogan\textsuperscript{1,2} \\
\textsuperscript{1}KUIS AI Center, Koc University, Turkey \quad \textsuperscript{2}EEE Department, Koc University, Turkey \\ \textsuperscript{3}CE Department, Koc University, Turkey \quad \textsuperscript{4}Google Cloud AI Research, Sunnyvale, CA\\
\texttt{\{sozsoy19, shamdan17, dyuret, alperdogan\}@ku.edu.tr}\quad 
\texttt{soarik@google.com}}

\begin{document}
\input{definitions.tex}

\maketitle

\begin{abstract}
Self-supervised learning allows AI systems to learn effective representations from large amounts of data using tasks that do not require costly labeling. 
Mode collapse, i.e., the model producing identical representations for all inputs, is a central problem to many self-supervised learning approaches, making self-supervised tasks, such as matching distorted variants of the inputs, ineffective.
In this article, we argue that a straightforward application of information maximization among alternative latent representations of the same input naturally solves the collapse problem and achieves competitive empirical results. We propose a self-supervised learning method, CorInfoMax, that uses a second-order statistics-based mutual information measure that reflects the level of correlation among its arguments. Maximizing this correlative information measure between alternative representations of the same input serves two purposes: (1) it avoids the collapse problem by generating feature vectors with non-degenerate covariances; (2) it establishes relevance among alternative representations by increasing the linear dependence among them. An approximation of the proposed information maximization objective simplifies to a Euclidean distance-based objective function regularized by the log-determinant of the feature covariance matrix. The regularization term acts as a natural barrier against feature space degeneracy. Consequently, beyond avoiding complete output collapse to a single point, the proposed approach also prevents dimensional collapse by encouraging the spread of information across the whole feature space. 
Numerical experiments demonstrate that CorInfoMax achieves better or competitive performance results relative to the state-of-the-art SSL approaches.

\end{abstract}

\section{Introduction}
\label{sec:introduction}
\input{introductionnewreb.tex}

\section{CorInfoMax as a criterion based on log-determinant mutual information}
\label{sec:ldmi}

\input{ldmireb.tex}

\section{LDMI based self-supervised learning}
\label{sec:ldmiss}

\input{ldmibasedbssreb.tex}

\section{Experiments}
\label{sec:experiments}
\input{experimentsreb2.tex}

\section{Discussions and conclusions}
\label{sec:conclusion}
We proposed a novel SSL framework based on the correlative information maximization, CorInfoMax, which provides a natural solution for obtaining organized representations that are scattered in the latent space and avoiding collapse. Our experiments demonstrate the state-of-the-art performance of CorInfoMax in numerous downstream tasks. The loss function of CorInfoMax does not have any significant impact on training time. 
Future work can further improve CorInfoMax focusing on shortcomings like improved selection of hyperparameter and augmentation, a core problem in SSL overall.

\begin{ack}
This work/research was supported by KUIS AI Center Research Award. We gratefully acknowledge the support of Google Cloud Credit Award.
\end{ack}

\medskip

\bibliography{articleneurips}

%%%%%%%%%%%%%%%%%%%%%%%%%%%%%%%%%%%%%%%%%%%%%%%%%%%%%%%%%%%%
\section*{Checklist}

%%% BEGIN INSTRUCTIONS %%%
%The checklist follows the references.  Please
%read the checklist guidelines carefully for information on how to answer these
%questions.  For each question, change the default \answerTODO{} to \answerYes{},
%\answerNo{}, or \answerNA{}.  You are strongly encouraged to include a {\bf
%justification to your answer}, either by referencing the appropriate section of
%your paper or providing a brief inline description.  For example:
%\begin{itemize}
%  \item Did you include the license to the code and datasets? \answerYes{See Section~\ref{gen_inst}.}
%  \item Did you include the license to the code and datasets? \answerNo{The code and the data are proprietary.}
%  \item Did you include the license to the code and datasets? \answerNA{}
%\end{itemize}
%Please do not modify the questions and only use the provided macros for your
%answers.  Note that the Checklist section does not count towards the page
%imit.  In your paper, please delete this instructions block and only keep the
%Checklist section heading above along with the questions/answers below.
%%% END INSTRUCTIONS %%%

\begin{enumerate}

\item For all authors...
\begin{enumerate}
  \item Do the main claims made in the abstract and introduction accurately reflect the paper's contributions and scope?
    \answerYes{As claimed in the abstract and introduction, in the article we provide a novel self supervised learning framework based on the information maximization principle and provide numerical experiment results reflecting its performance.}
  \item Did you describe the limitations of your work?
    \answerYes{We stated sensitivity to hyperparameter selections as a limitation in Section \ref{sec:conclusion}. }
  \item Did you discuss any potential negative societal impacts of your work?
    \answerNA{We  believe that our work does not have any potential negative societal impacts.}
  \item Have you read the ethics review guidelines and ensured that your paper conforms to them?
     \answerYes{We have read the ethics review guidelines and our paper conforms to them.}
\end{enumerate}

\item If you are including theoretical results...
\begin{enumerate}
  \item Did you state the full set of assumptions of all theoretical results?
     \answerYes{Theoretical discussions in Section \ref{sec:ldmi} and Section \ref{sec:ldmiss} clearly provide the mathematical settings used in our derivations.  }
        \item Did you include complete proofs of all theoretical results?
    \answerYes{We show the details of the derivations of our proposed objective in Section \ref{sec:ldmi}.}
\end{enumerate}

\item If you ran experiments...
\begin{enumerate}
  \item Did you include the code, data, and instructions needed to reproduce the main experimental results (either in the supplemental material or as a URL)?
    \answerYes{We include the Python codes of our experiments in supplementary materials and the details of the experiment results in Section \ref{sec:experiments} and in Appendix.}
  \item Did you specify all the training details (e.g., data splits, hyperparameters, how they were chosen)?
    \answerYes{We provided all the details about training, its phases, hyperparameter selections in Section \ref{sec:experiments} and in Appendix.}
        \item Did you report error bars (e.g., with respect to the random seed after running experiments multiple times)?
    \answerYes{In Appendix, we provide information on the confidence interval of our experimental results.}
        \item Did you include the total amount of compute and the type of resources used (e.g., type of GPUs, internal cluster, or cloud provider)?
    \answerYes{We reported the details about the computation resources and computation times in Section \ref{sec:experiments} and in Appendix.}
\end{enumerate}

\item If you are using existing assets (e.g., code, data, models) or curating/releasing new assets...
\begin{enumerate}
  \item If your work uses existing assets, did you cite the creators?
    \answerYes{We cite references for all the data sets, experiment results and network models in Section \ref{sec:experiments}. }
  \item Did you mention the license of the assets?
    \answerNA{}
  \item Did you include any new assets either in the supplemental material or as a URL?
    \answerYes{We include our code as supplementary material.}
  \item Did you discuss whether and how consent was obtained from people whose data you're using/curating?
    \answerNA{}
  \item Did you discuss whether the data you are using/curating contains personally identifiable information or offensive content?
    \answerNA{}
\end{enumerate}

\item If you used crowdsourcing or conducted research with human subjects...
\begin{enumerate}
  \item Did you include the full text of instructions given to participants and screenshots, if applicable?
    \answerNA{}
  \item Did you describe any potential participant risks, with links to Institutional Review Board (IRB) approvals, if applicable?
    \answerNA{}
  \item Did you include the estimated hourly wage paid to participants and the total amount spent on participant compensation?
 \answerNA{}
\end{enumerate}

\end{enumerate}

%%%%%%%%%%%%%%%%%%%%%%%%%%%%%%%%%%%%%%%%%%%%%%%%%%%%%%%%%%%%

\appendix

%\section{Appendix}
\input{appendixreb2.tex}

\end{document}

%% file: definitions.tex
\newcommand{\ba}{\mathbf{a}}
\newcommand{\vb}{\mathbf{b}}
\newcommand{\be}{\mathbf{e}}
\newcommand{\bx}{\mathbf{x}}
\newcommand{\bh}{\mathbf{h}}
\newcommand{\bu}{\mathbf{u}}
\newcommand{\bv}{\mathbf{v}}
\newcommand{\by}{\mathbf{y}}
\newcommand{\bq}{\mathbf{q}}
\newcommand{\bg}{\mathbf{g}}
\newcommand{\bY}{\mathbf{Y}}
\newcommand{\bX}{\mathbf{X}}
\newcommand{\bU}{\mathbf{U}}
\newcommand{\bV}{\mathbf{V}}
\newcommand{\bB}{\mathbf{B}}
\newcommand{\bR}{\mathbf{R}}
\newcommand{\bs}{\mathbf{s}}
\newcommand{\bz}{\mathbf{z}}
\newcommand{\bS}{\mathbf{S}}
\newcommand{\bL}{\mathbf{L}}
\newcommand{\bK}{\mathbf{K}}
\newcommand{\bH}{\mathbf{H}}
\newcommand{\bZ}{\mathbf{Z}}
\newcommand{\vB}{\mathbf{B}}
\newcommand{\bw}{\mathbf{w}}
\newcommand{\bW}{\mathbf{W}}
\newcommand{\bA}{\mathbf{A}}
\newcommand{\bC}{\mathbf{C}}
\newcommand{\bI}{\mathbf{I}}
\newcommand{\bF}{\mathbf{F}}
\newcommand{\bQ}{\mathbf{Q}}
\newcommand{\Ball}{\mathcal{B}}
\newcommand{\Tin}{\mathcal{T}_{in}}
\newcommand{\Tout}{\mathcal{T}_{out}}
\newcommand{\Pcal}{\mathcal{P}}
\newcommand{\Ycal}{\mathcal{Y}}
\newcommand{\Scal}{\mathcal{S}}
\newcommand{\Pex}{\Pcal_{\text{ex}}}
\newcommand{\bD}{\mathbf{D}}
\newcommand{\bPi}{\bm{\Pi}}
\newcommand{\bmu}{\boldsymbol{\mu}}
\newcommand{\blambda}{\bm{\lambda}}
\newcommand{\bSigma}{\bm{\Sigma}}
\newcommand{\bGam}{\bm{\Gamma}}
\newcommand{\cvert}{\hspace{0.02in} \vert \hspace{0.02in}}
\newcommand{\hld}{h_{LD}^{(\varepsilon)}}
\newcommand{\ild}{I_{LD}^{(\varepsilon)}}
\newtheorem{lemma}{Lemma}
\newtheorem{definition}{Definition}

% BLOCK DIAGRAM RELATED
% Color Definitions
\definecolor{darkblue}{HTML}{1f4e79}
\definecolor{lightblue}{HTML}{00b0f0}
\definecolor{salmon}{HTML}{ff9c6b}
\definecolor{bluegray}{rgb}{0.4, 0.6, 0.8}
\definecolor{junglegreen}{rgb}{0.16, 0.67, 0.53}

\definecolor{coldgreen}{RGB}{231,249,223}
\definecolor{coldgreen2}{RGB}{188,218,196}
\definecolor{coldblue}{RGB}{219,240,254}
\definecolor{coldorange}{RGB}{248,216,167}
\definecolor{coldyellow}{RGB}{248,247,201}
\definecolor{coldpurple}{RGB}{201,210,248}
\definecolor{coldred}{RGB}{251,169,169}
\definecolor{coldgray}{RGB}{221,218,217}

% Define parallelepiped shape:
\makeatletter
\pgfkeys{/pgf/.cd,
  parallelepiped offset x/.initial=2mm,
  parallelepiped offset y/.initial=2mm
}
\pgfdeclareshape{parallelepiped}
{
  \inheritsavedanchors[from=rectangle] % this is nearly a rectangle
  \inheritanchorborder[from=rectangle]
  \inheritanchor[from=rectangle]{north}
  \inheritanchor[from=rectangle]{north west}
  \inheritanchor[from=rectangle]{north east}
  \inheritanchor[from=rectangle]{center}
  \inheritanchor[from=rectangle]{west}
  \inheritanchor[from=rectangle]{east}
  \inheritanchor[from=rectangle]{mid}
  \inheritanchor[from=rectangle]{mid west}
  \inheritanchor[from=rectangle]{mid east}
  \inheritanchor[from=rectangle]{base}
  \inheritanchor[from=rectangle]{base west}
  \inheritanchor[from=rectangle]{base east}
  \inheritanchor[from=rectangle]{south}
  \inheritanchor[from=rectangle]{south west}
  \inheritanchor[from=rectangle]{south east}
  \backgroundpath{
    % store lower right in xa/ya and upper right in xb/yb
    \southwest \pgf@xa=\pgf@x \pgf@ya=\pgf@y
    \northeast \pgf@xb=\pgf@x \pgf@yb=\pgf@y
    \pgfmathsetlength\pgfutil@tempdima{\pgfkeysvalueof{/pgf/parallelepiped
      offset x}}
    \pgfmathsetlength\pgfutil@tempdimb{\pgfkeysvalueof{/pgf/parallelepiped
      offset y}}
    \def\ppd@offset{\pgfpoint{\pgfutil@tempdima}{\pgfutil@tempdimb}}
    \pgfpathmoveto{\pgfqpoint{\pgf@xa}{\pgf@ya}}
    \pgfpathlineto{\pgfqpoint{\pgf@xb}{\pgf@ya}}
    \pgfpathlineto{\pgfqpoint{\pgf@xb}{\pgf@yb}}
    \pgfpathlineto{\pgfqpoint{\pgf@xa}{\pgf@yb}}
    \pgfpathclose
    \pgfpathmoveto{\pgfqpoint{\pgf@xb}{\pgf@ya}}
    \pgfpathlineto{\pgfpointadd{\pgfpoint{\pgf@xb}{\pgf@ya}}{\ppd@offset}}
    \pgfpathlineto{\pgfpointadd{\pgfpoint{\pgf@xb}{\pgf@yb}}{\ppd@offset}}
    \pgfpathlineto{\pgfpointadd{\pgfpoint{\pgf@xa}{\pgf@yb}}{\ppd@offset}}
    \pgfpathlineto{\pgfqpoint{\pgf@xa}{\pgf@yb}}
    \pgfpathmoveto{\pgfqpoint{\pgf@xb}{\pgf@yb}}
    \pgfpathlineto{\pgfpointadd{\pgfpoint{\pgf@xb}{\pgf@yb}}{\ppd@offset}}
  }
}
\makeatother

\tikzset{
  % Dark blue blocks
  block/.style={
    parallelepiped,fill=white, draw,
    minimum width=2.4cm,
    minimum height=1.3cm,
    parallelepiped offset x=0.5cm,
    parallelepiped offset y=0.5cm,
    path picture={
      \draw[top color=darkblue,bottom color=darkblue]
        (path picture bounding box.south west) rectangle 
        (path picture bounding box.north east);
    },
    text=white,
  },
  % Orange-ish blocks
  conv/.style={
    parallelepiped,fill=white, draw,
    minimum width=1.5cm,
    minimum height=2.4cm,
    parallelepiped offset x=0.5cm,
    parallelepiped offset y=0.5cm,
    path picture={
      \draw[top color=salmon,bottom color=salmon]
        (path picture bounding box.south west) rectangle 
        (path picture bounding box.north east);
    },
    text=white,
  },
    % bluegray-ish blocks
   inputb/.style={
    parallelepiped,fill=white, draw,
    minimum width=1.5cm,
    minimum height=2.4cm,
    parallelepiped offset x=0.5cm,
    parallelepiped offset y=0.5cm,
    path picture={
      \draw[top color=gray,bottom color=gray]
        (path picture bounding box.south west) rectangle 
        (path picture bounding box.north east);
    },
    text=white,
  },
     % salmon-ish blocks
   inputbA/.style={
    parallelepiped,fill=white, draw,
    minimum width=1.5cm,
    minimum height=2.4cm,
    parallelepiped offset x=0.5cm,
    parallelepiped offset y=0.5cm,
    path picture={
      \draw[top color=bluegray,bottom color=salmon]
        (path picture bounding box.south west) rectangle 
        (path picture bounding box.north east);
    },
        text=white,
  },
       % red-ish blocks
   inputbB/.style={
    parallelepiped,fill=white, draw,
    minimum width=1.5cm,
    minimum height=2.4cm,
    parallelepiped offset x=0.5cm,
    parallelepiped offset y=0.5cm,
    path picture={
      \draw[top color=bluegray,bottom color=red]
        (path picture bounding box.south west) rectangle 
        (path picture bounding box.north east);
    },
    text=white,
  },
      % junglegreen-ish blocks
   dnnb/.style={
    parallelepiped,fill=white, draw,
    minimum width=3.5cm,
    minimum height=2.4cm,
    parallelepiped offset x=0.5cm,
    parallelepiped offset y=0.5cm,
    path picture={
      \draw[top color=junglegreen,bottom color=junglegreen]
        (path picture bounding box.south west) rectangle 
        (path picture bounding box.north east);
    },
    text=white,
  },
  % Taller Light blue blocks:
  plate/.style={
    parallelepiped,fill=white, draw,
    minimum width=0.1cm,
    minimum height=7.4cm,
    parallelepiped offset x=0.5cm,
    parallelepiped offset y=0.5cm,
    path picture={
      \draw[top color=lightblue,bottom color=lightblue]
        (path picture bounding box.south west) rectangle 
        (path picture bounding box.north east);
    },
    text=white,
  },
  % Arrows between blocks:
  link/.style={
    color=lightblue,
    line width=2mm,
  },
      % new input blocks
   inputbnew/.style={
    parallelepiped,fill=white, draw,
    minimum width=1.5cm,
    minimum height=2.4cm,
    parallelepiped offset x=0.5cm,
    parallelepiped offset y=0.5cm,
    path picture={
      \draw[top color=coldgray,bottom color=coldgray]
        (path picture bounding box.south west) rectangle 
        (path picture bounding box.north east);
    },
    text=black,
  },
       % new input blocks 2
   inputbAnew/.style={
    parallelepiped,fill=white, draw,
    minimum width=1.5cm,
    minimum height=2.4cm,
    parallelepiped offset x=0.5cm,
    parallelepiped offset y=0.5cm,
    path picture={
      \draw[top color=coldgreen2,bottom color=coldgreen2]
        (path picture bounding box.south west) rectangle 
        (path picture bounding box.north east);
    },
        text=black,
  },
       % new input blocks 3
   inputbBnew/.style={
    parallelepiped,fill=white, draw,
    minimum width=1.5cm,
    minimum height=2.4cm,
    parallelepiped offset x=0.5cm,
    parallelepiped offset y=0.5cm,
    path picture={
      \draw[top color=coldorange,bottom color=coldorange]
        (path picture bounding box.south west) rectangle 
        (path picture bounding box.north east);
    },
    text=black,
  },
}

\pgfarrowsdeclarecombine{|<}{>|}{|}{|}{latex}{latex}
\def\Dimline[#1][#2][#3]{
    %\node at (0,0) {"test: #1 - #2 ..."};
    \begin{scope}[>=latex] % redef arrow for dimension lines
        \draw[|<->|,
        decoration={markings, % switch on markings
                mark=at position .5 with {\node[gray] at (0,0.25) {\tiny{#3}};},
        },
        postaction=decorate] #1 -- #2 ;
    \end{scope}
}

\def\DimlineN[#1][#2][#3]{
    %\node at (0,0) {"test: #1 - #2 ..."};
    \begin{scope}[>=latex] % redef arrow for dimension lines
        \draw[|<->|,
        decoration={markings, % switch on markings
                mark=at position .5 with {\node[gray,rotate=90] at (0,-0.25) {\tiny{#3}};},
        },
        postaction=decorate] #1 -- #2 ;
    \end{scope}
}
\def\DimlineK[#1][#2][#3]{
    %\node at (0,0) {"test: #1 - #2 ..."};
    \begin{scope}[>=latex] % redef arrow for dimension lines
        \draw[|<->|,
        decoration={markings, % switch on markings
                mark=at position .5 with {\node[gray] at (0,-0.19) {\tiny{#3}};},
        },
        postaction=decorate] #1 -- #2 ;
    \end{scope}
}
\def\DimlineP[#1][#2][#3]{
    %\node at (0,0) {"test: #1 - #2 ..."};
    \begin{scope}[>=latex] % redef arrow for dimension lines
        \draw[|<->|,
        decoration={markings, % switch on markings
                mark=at position .5 with {\node[gray,rotate=90] at (0,0.25) {\tiny{#3}};},
        },
        postaction=decorate] #1 -- #2 ;
    \end{scope}
}

\def\DimlineKnew[#1][#2][#3]{
    %\node at (0,0) {"test: #1 - #2 ..."};
    \begin{scope}[>=latex] % redef arrow for dimension lines
        \draw[|<->|,
        decoration={markings, % switch on markings
                mark=at position .5 with {\node[gray,rotate=-45] at (0.1,-0.25) {#3};},
        },
        postaction=decorate] #1 -- #2 ;
    \end{scope}
}

\def\DimlineKneww[#1][#2][#3]{
    %\node at (0,0) {"test: #1 - #2 ..."};
    \begin{scope}[>=latex] % redef arrow for dimension lines
        \draw[|<->|,
        decoration={markings, % switch on markings
                mark=at position .5 with {\node[gray,rotate=-45] at (-0.1,-0.25) {#3};},
        },
        postaction=decorate] #1 -- #2 ;
    \end{scope}
}

\definecolor{commentcolor}{RGB}{110,154,155}   % define comment color
\newcommand{\PyComment}[1]{\ttfamily\textcolor{commentcolor}{\# #1}}  % add a "#" before the input text "#1"
\newcommand{\PyCode}[1]{\ttfamily\textcolor{black}{#1}} % \ttfamily is the code font

%% file: introductionnewreb.tex
%%%%%%%%%%%%%%%%%%%%%%%%%%%%%%%%%%%%%%%%%%%%%%%%%%%%%%%%%%
%% Description and benefits of self-supervised learning %%
%%%%%%%%%%%%%%%%%%%%%%%%%%%%%%%%%%%%%%%%%%%%%%%%%%%%%%%%%%

\newif\ifanswers

Self-supervised learning (SSL) is an important paradigm with significant impact in artificial intelligence \citep{collobert2008unified, doersch2015unsupervised, devlin2018bert,grill2020bootstrap,zbontar2021barlow,bardes2021vicreg}. In particular, SSL can significantly reduce the need for expensively-obtained labeled data. Furthermore, the internal representations learned by SSL approaches are more suitable for task generalization in transfer learning applications \citep{ericsson2021well}. Recently, SSL methods have even been demonstrated to achieve comparable performance to their supervised counterparts \citep{chen2020simple,caron2020unsupervised}.

\begin{figure}[t]
\centering
\includegraphics[width=13.cm, trim=5cm 10.5cm 0.5cm 7.5cm,clip]{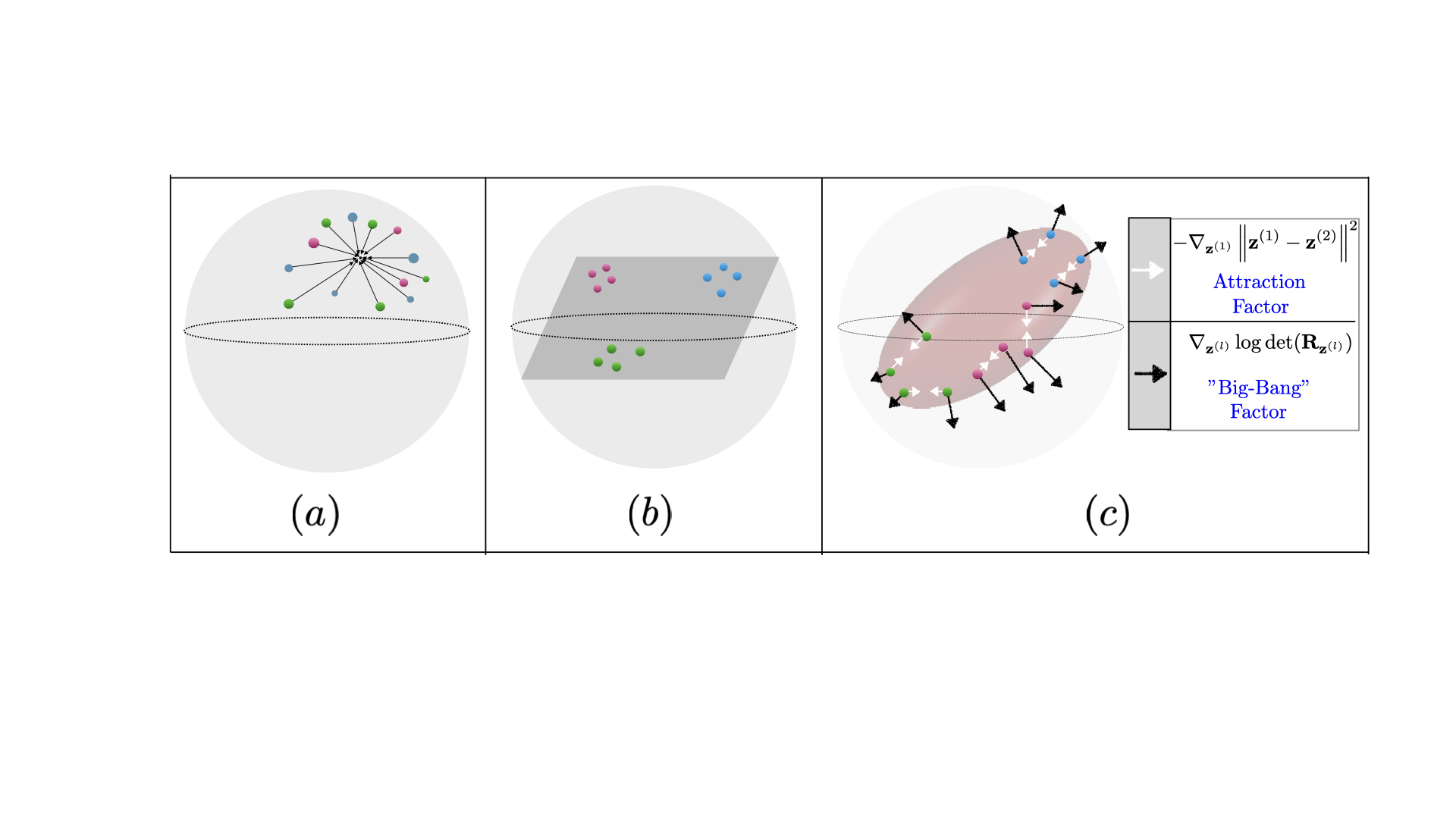}
\caption{Depictions in latent space - (a) "Total Output Collapse": all latent vectors converge to the same point, (b) "Dimensional Output Collapse": latent vectors are restricted to a strict subspace of the latent space, (c) Gradient dynamics of the CorInfoMax based on (\ref{eq:covdet2}): the ellipsoid surface reflects the average spread pattern of latent vectors. }
\label{fig:toypicture}
\end{figure}

\looseness=-1 The primary approach for SSL is to define a pretext objective based on unlabeled data, whose optimization leads to rich and useful representations for downstream applications. For example, for visual processing, a common way to construct a pretext task is to obtain similar features for two different augmentations of the same input, which exploits the expected invariance of representations to particular transformations of the input. However, the corresponding optimization setting should avoid the (total) ``collapse problem,'' which is described as generating the same feature output for all inputs.  Figure \ref{fig:toypicture}.(a) illustrates a toy view of the total collapse problem, where all latent vector outputs converge to a single point during training. We can also define its less severe version, which is referred to as "dimensional collapse," where all latent vectors lie in a strict subspace of the whole latent space as illustrated by Figure \ref{fig:toypicture}.(b).

This article proposes an SSL approach based on the maximization of a form of  mutual information among the alternative latent representations obtained from the same input. The mutual information maximization approach  would serve two fundamental purposes for SSL: 
\begin{itemize}
\item[(i).]({\it Similarity}) It would enforce the representations obtained for the same input source to be related to or dependent on each other,
\item[(ii).]({\it No Collapse}) Since the entropy of each alternative representation is lower bounded by the mutual information, maximizing mutual information ensures that latent representations have non-degenerate distributions avoiding collapse.
\end{itemize}
Therefore, the information maximization approach provides a  natural solution to the problems targeted by the existing SSL approaches, which are briefly surveyed in Sec. \ref{sec:ssllit}.

The conventional choice for mutual information measure is Shannon Mutual Information (SMI) \citep{thomas2006elements}. 
For SSL, the fundamental limitations on the precise computation of SMI \citep{poole2019variational,tschannen2019mutual, mcallester2020formal} impose serious challenges on model training. Especially when mutual information is large, we require drastically more samples (exponentially related to SMI) to obtain a reliable estimate of SMI \cite{mcallester2020formal}. However, the use of large batch sizes to improve precision may not be a favorable solution, as (i) it may have an adverse impact on generalization performance \citep{mccandlish2018empirical,masters2018revisiting}, (ii)  it can significantly increase memory requirements. Furthermore,   estimating and optimizing SMI can significantly increase the computational burden. Finally, SMI maximization may induce a nonlinear dependence between the representations corresponding to the same input, where the resultant partitioning of the latent space may not be suitable as an input to a simple linear classifier. 

To address these issues, we can consider a second-order-statistics-based mutual information measure, referred to as log-determinant mutual information (LDMI) \citep{zhouyin:21,erdogan2022blind}. The joint entropy measure corresponding to LDMI is the log-determinant of the diagonally perturbed covariance matrix. The zero perturbation case boils down to the Shannon differential entropy for Gaussian distributed vectors. Unlike SMI, which reflects general dependence between its arguments, LDMI reflects linear dependence and is computationally more efficient.

In this article, we propose a novel SSL approach, correlative information maximization (CorInfoMax), derived from  the LDMI measure. The loss function for CorInfoMax is obtained through the first-order approximation of LDMI and  restricting the linear dependence to the identity map. The resulting optimization objective is  a Euclidean distance-based loss function regularized by the log-determinant of the latent vector covariance matrix. This regularization term encourages the spreading of the feature vectors and acts as a natural barrier against the dimensional degeneracy and collapse in the feature space.  Our numerical experiments confirm that CorInfoMax learns effective representations that perform well in downstream tasks. 

The main contributions can be summarized as follows:
\begin{itemize}
    \item Based on information theoretical grounds, we propose a novel framework, CorInfoMax, with an interpretable loss function that has explicit components for minimizing variance of positive samples (`the attraction factor') and for making full use of the embedding space to avoid dimensional collapse (`the big-bang factor').  
    \item We introduce a computationally-efficient approximation to LDMI that simplifies it by using an identity mapping instead of a general linear mapping.  This approximation directly minimizes representation variance, which is desired for self-supervised learning.
    \item CorInfoMax relies only on second-order statistics, does not require negative samples, and does not force embeddings to be uncorrelated.
    \item In addition to these theoretically appealing features, the proposed CorInfoMax framework achieves either better or similar performance results compared to the state-of-the-art SSL approaches.  
\end{itemize}

The following is the organization of the article: Section \ref{sec:relatedwork} provides a discussion of the relevant literature related to the proposed approach. We introduce  the log-determinant entropy and mutual information measures by highlighting LDMI as a measure of  linear dependence in Section \ref{sec:ldmi}. In the same section, we derive the objective function of CorInfoMax as a variation on the LDMI measure. Section \ref{sec:ldmiss} introduces the proposed correlative information maximization-based  self-supervised learning method. Section \ref{sec:experiments} provides the numerical experiments illustrating the performance of the proposed approach. The appendix provided in the supplementary document contains the details related to these experiments.  Finally, Section \ref{sec:conclusion} presents the discussion and conclusions. %\dy{We need a conclusion/contributions section at the end!}

%%%%%%%%%%%%%%%%%%%%%%%%%%%%%%%%%%%%%%%%%%%%%%%%%%%%%%%%%%%%%%%%%
%% Underlining novelty and advantages of the proposed approach %%
%%%%%%%%%%%%%%%%%%%%%%%%%%%%%%%%%%%%%%%%%%%%%%%%%%%%%%%%%%%%%%%%%
% - principled approach
% - based on information theory
% - superior/comparable performance to the state of the art
% - nicely interpretable optimization objective

 \section{Related work}
 \label{sec:relatedwork}
 \subsection{Determinant maximization for unsupervised learning}
 \label{sec:detmaxlit}
 The determinant maximization criterion utilized in our framework has been used as an effective algorithmic tool in unsupervised matrix factorization methods such as nonnegative matrix factorization (NMF) \citep{fu2019nonnegative, fu2016robust}, simplex-structured matrix factorization (SSMF) \citep{chan2011simplex}, sparse component analysis (SCA) \citep{babatas2018algorithmic}, bounded component analysis (BCA) \citep{inan2014convolutive} and polytopic matrix factorization (PMF) \citep{tatli:21tsp}.  
 
\looseness=-1 In the generative models of these frameworks, the input data is assumed to be linear transformations of some latent vectors. Furthermore, these latent vectors are assumed to be sufficiently scattered in their  domain.  Maximizing the determinant of the latent covariance matrix spreads the latent vector estimates to capture this presumed scattering of the generative model samples.  
Similarly, the log-determinant of the latent vector covariance matrix in the CorInfoMax objective causes the spreading of latent vectors in their ambient space  and,  therefore, avoids collapse. According to \citep{erdogan2022blind}, the determinant maximization criterion used in these matrix factorization frameworks is essentially  equivalent to correlative information maximization between  input and its latent factors based on the LDMI objective, where the latent vectors  are constrained to lie in their domain determined by the generative model. 

 In the CorInfoMax approach, we utilize determinant maximization for correlative information maximization among latent space vectors rather than between inputs and their latent space representations.
 
\subsection{Handling collapse for SSL}
\label{sec:ssllit}
\looseness=-1 Collapse is a central concern for SSL with the same input in computer vision, and we can categorize the existing approaches according to how they handle this issue. For example, contrastive methods, such as SimCLR \citep{chen2020simple} and MoCo  \citep{he2020momentum}, define an objective that pushes features for different inputs (negative samples) away from each other while keeping the representations corresponding to the same input (positive samples) close to each other. The number and the choice of the negative samples appear as critical factors for the performance and scalability of contrastive methods. Another category of SSL is the distillation methods, such as BYOL \citep{grill2020bootstrap} and SimSiam \citep{chen2021exploring}, which avoid collapse through the asymmetry of alternative encoder branches  and algorithmic tuning \citep{tian2021understanding}. Other SSL approaches such as DeepCluster \citep{caron2018deep}, SeLa \citep{asano2019self} and SvAW \citep{caron2020unsupervised} enforce a cluster structure in the feature space to prevent constant output. As a different line of algorithms, decorrelated batch normalization (DBN) \citep{huang2018decorrelated} Barlow-Twins \citep{zbontar2021barlow}, Whitening MSE (W-MSE) \citep{ermolov2021whitening}  and VICReg \citep{bardes2021vicreg}, use feature decorrelation \citep{hua2021feature} as a means to avoid information collapse. The effective whitening mechanisms used in these methods aim for the isotropic spread of information inside the feature space which also prevents "dimensional collapse" illustrated in Figure \ref{fig:toypicture}.(b). The reference \cite{li2021self} constructs an SSL loss function using the Hilbert-Schmidt Independence Criterion, a kernel-based independence measure, to maximize the dependence between alternative representations of the same input. Finally, \citep{haochen2021provable} proposed graph representations for the embeddings of positive samples and the corresponding spectral decomposition algorithm.   
 
The proposed CorInfoMax approach is most related to the decorrelation-based methods discussed above, mainly due to correlation-based measures.  However, unlike the decorrelation methods, CorInfoMax does not constrain latent vectors to be uncorrelated. Instead, it avoids covariance matrix degeneracy by using its log-determinant as a regularizer loss function.  Furthermore, the information maximization principle is more direct and explicit for the CorInfoMax algorithm. 
 
\subsection{Information maximization for unsupervised learning}
\label{sec:infomaxlit}
The maximization of SMI has been proposed as an unsupervised learning mechanism in different but related contexts. As one of the earliest approaches, Linsker proposed maximum information transfer from input data to its latent representation and showed that it is equivalent to maximizing the determinant of the output covariance under the Gaussian distribution assumption  \citep{linsker1988self}.  Around the same time frame, Becker \& Hinton  \citep{becker1992self} put forward a representation learning approach based on the maximization of (an approximation of) the SMI between the alternative latent vectors obtained from the same image.   The most well-known application is the Independent Component Analysis (ICA) Infomax algorithm \citep{bell1995information} for separating independent sources from their linear combinations. The ICA-Infomax algorithm targets to maximize the mutual information between mixtures and source estimates while imposing statistical  independence among outputs. 
The Deep Infomax approach \citep{hjelm2018learning} extends this idea to unsupervised feature learning by maximizing the mutual information between the input and output while matching a prior distribution for the representations.  

It is essential to underline that the proposed method clearly distinguishes itself from the Deep Infomax in \citep{hjelm2018learning}: Our objective is not to maximize the mutual information between inputs and outputs of a deep network. Instead, we maximize the mutual information content of the alternative latent representations of the same input. From this point of view, our approach is closer to what is aimed at by \citep{becker1992self}. However, we use a different (correlative) information measure, which is computationally more efficient and induces a special form of linear dependence among alternative latent representations of the same input, which may be more desirable considering the goal of generating features for a linear classifier.

%\subsection{Organization of the Article}
%%%%%%%%%%%%%%%%%%%%%%%%%%%%%%%%%%%%%%%%%%%%%%%%%%%%%%%%%%%%%%%%%
%% Describe the story-line of the article                      %%
%%%%%%%%%%%%%%%%%%%%%%%%%%%%%%%%%%%%%%%%%%%%%%%%%%%%%%%%%%%%%%%%%

%% file: ldmireb.tex
\looseness=-1 This section derives the optimization objective for the  CorInfoMax framework as a variation on the LDMI measure. As described in Appendix \ref{app:infomeas}, the LDMI between random vectors $\bx$ and $\by$ is given by
\begin{eqnarray}
\ild(\bx;\by)&=&\frac{1}{4}(\log\det(\bR_\bx+\varepsilon \mathbf{I})+\log\det(\bR_\by+\varepsilon \mathbf{I}) \nonumber \\&-&\log\det(\bR_\bx-\bR_{\bx\by}(\bR_\by+\varepsilon\mathbf{I})^{-1}\bR_{\bx\by}^T+\varepsilon\mathbf{I})\nonumber \\
 &-&\log\det(\bR_\by-\bR_{\bx\by}^T(\bR_\bx+\varepsilon\mathbf{I})^{-1}\bR_{\bx\by}+\varepsilon\mathbf{I})), \label{eq:ldmi3}
\end{eqnarray}
where $\bR_\bx$ and $\bR_\by$ are auto-covariance matrices for $\bx$ and $\by$, respectively, and $\bR_{\bx\by}$ is their cross-covariance matrix.

There are three potential advantages of using LDMI in (\ref{eq:ldmi3}) over SMI in (\ref{eq:smi}) for SSL:
\begin{itemize}
    \item[i.] It is only based on  second-order statistics. On the other hand, SMI is a statistic based on the joint PDFs of the argument vectors, and its accurate estimation has high sample complexity. In contrast, it is more practical to obtain (the estimates of) the auto-covariance and the cross-covariance matrices, as discussed in Section \ref{sec:ldmiss}.
    \item[ii.] There are benefits of information maximization principle in connection with self-supervised training. The maximization of SMI of two vectors induces a general, potentially nonlinear dependence between them, whereas the maximization of LDMI increases correlation, or linear dependence. Therefore, LDMI-based information maximization is expected to organize the feature space being more favorable for data-efficient and low-complexity linear or shallow supervised classifiers as targeted in SSL applications.
    \item[iii.] The resulting LDMI-based objective functions are more interpretable -- they involve the log-determinant of the projector-space covariance, whose maximization clearly avoids feature collapse, a significant concern in SSL methods.% \dy{how is avoiding collapse related to interpretability?}
\end{itemize}

The primary motivation for obtaining an LDMI variant is to increase the similarity between the alternative latent representations of the same input by enforcing the identity transformation between them. 

\looseness=-1 Let $\bz^{(1)}$ and $\bz^{(2)}$ represent alternative latent representations corresponding to the same input. One potential SSL approach would be to pursue direct maximization of $\ild(\bz^{(1)};\bz^{(2)})$ in  (\ref{eq:ldmi3}). As discussed earlier, this would maximize the correlation between $\bz^{(1)}$ and $\bz^{(2)}$, therefore inducing a linear dependence between them. For the SSL application, we would prefer the identity mapping to an arbitrary linear relationship such that the alternative latent representations corresponding to the same input concentrate in the same neighborhood of the latent space. Therefore,  we modify the LDMI expression in (\ref{eq:ldmi3}) to impose this constraint.  
For this purpose, we apply the first-order Taylor series approximation, $\log\det(\bC+\bD)\approx\log\det(\bC)+\textrm{Tr}(\bD^T\bC^{-1})$, on the third term in the right side of (\ref{eq:ldmi3}), which provides
\begin{eqnarray}
&&\log\det({\bR}_{\bz^{(1)}} -{\bR}_{\bz^{(1)}\bz^{(2)}}({\bR}_{\bz^{(2)}}+\varepsilon\mathbf{I})^{-1}{\bR}_{\bz^{(1)}\bz^{(2)}}^T+\varepsilon\mathbf{I})
\nonumber \\
&&\approx
\frac{1}{\varepsilon}\textrm{Tr}({\bR}_{\bz^{(1)}} -{\bR}_{\bz^{(1)}\bz^{(2)}}{\bR}^{-1}_{\bz^{(2)}}{\bR}_{\bz^{(1)}\bz^{(2)}}^T)\nonumber+\log\det(\varepsilon \mathbf{I}) \\
&&=\frac{1}{\varepsilon}\min_{\bA_1,\vb_1}E(\|\bz^{(1)}-(\bA_1\bz^{(2)}+\vb_1)\|_2^2)+P\log(\varepsilon). \label{eq:mseaffine}
\end{eqnarray}
Therefore, the expression in (\ref{eq:mseaffine}) corresponds to the mean square error of the best linear (affine) MMSE estimator of $\bz^{(1)}$ from $\bz^{(2)}$, multiplied by $\varepsilon^{-1}$.  Similarly, if we apply the same approximation to the fourth term on the right side of (\ref{eq:ldmi3}), we obtain 
\begin{eqnarray}
\log\det({\bR}_{\bz^{(2)}} -{\bR}_{\bz^{(1)}\bz^{(2)}}^T({\bR}_{\bz^{(1)}}+\varepsilon\mathbf{I})^{-1}{\bR}_{\bz^{(1)}\bz^{(2)}}+\varepsilon\mathbf{I})\nonumber \\
\approx
\frac{1}{\varepsilon}\min_{\bA_2,\vb_2}E(\|\bz^{(2)}-(\bA_2\bz^{(1)}+\vb_2)\|_2^2)+P\log(\varepsilon). \label{eq:mseaffine2}
\end{eqnarray}
 \looseness=-1 In order to induce the identity mapping between $\bz^{(1)}$ and $\bz^{(2)}$, we constrain $\bA_1=\bA_2=\mathbf{I}$, and $\vb_1=\vb_2=\mathbf{0}$, which transforms both (\ref{eq:mseaffine}) and ((\ref{eq:mseaffine2}) into $\varepsilon^{-1}E(\|\bz^{(1)}-\bz^{(2)}\|_2^2)+\mbox{const}$. Therefore, scaling the expression in (\ref{eq:ldmi3}) by $4$  and using this modification, we obtain
\begin{eqnarray}
J(\bz^{(1)},\bz^{(2)})=\log\det({\bR}_{\bz^{(1)}}+\varepsilon \mathbf{I})+\log\det({\bR}_{\bz^{(2)}}+\varepsilon \mathbf{I})-2\varepsilon^{-1}E(\|\bz^{(1)}-\bz^{(2)}\|_2^2), \label{eq:stochcorinfomax}
\end{eqnarray}
 where we ignored the constant terms. We refer to (\ref{eq:stochcorinfomax}) as the stochastic CorInfoMax objective function. In Section \ref{sec:ldmiss}, we propose a SSL method based on the optimization of this objective function.

%% file: ldmibasedbssreb.tex
This section introduces the proposed correlative information maximization (CorInfoMax) approach for SSL.  
We start by describing the presumed setting for the pretext task, which is matching the latent representations of different augmentations of the same input in Section \ref{sec:setup}. Section \ref{sec:corInfoMax} is the main section where we propose the correlative information maximization algorithm for SSL.  Finally, we discuss the implementation complexity of the proposed approach in Section \ref{sec:complexity}. 

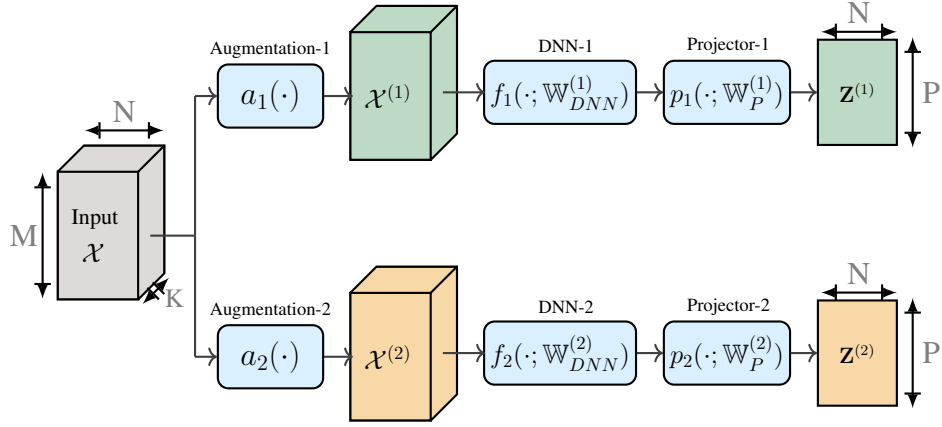
\begin{figure}[h]
\hspace*{-0.2in}\input{blockdiagram2branchv2.tex}
\caption{SSL setup: we consider two parallel encoder branches corresponding to two different augmentations of the same input $X$. Augmented views are fed into Siamese networks $f$ followed by projector $p$, basically a 3-layer MLP.  $N$ stands for batch size, $M \times K$ for image dimension, $P$ for feature dimension of projector output. }
\label{fig:ssblockdiagram}
\end{figure}

\subsection{Self-supervised learning setting}
\label{sec:setup}
We start by describing the presumed self-supervised learning setup, which is illustrated  in Figure \ref{fig:ssblockdiagram}:
\begin{itemize}
    \item \underline{\it The input} is a sequence of tensors $\{\mathcal{X}[l]\in\mathbb{R}^{M\times K\times N}, l \in \mathbb{Z}\}$, where each index sample corresponds to a batch of $N$ images with dimensions $M \times K$.%
    As discussed in Section \ref{sec:introduction}, the pretext task that we consider is matching the latent representations of the two augmentations of the same input. Therefore, we first apply \underline{\it the augmentation functions} $a_1(\cdot)$ and $a_2(\cdot)$ to obtain the transformed versions of the input sequence, i.e., 
    %\begin{eqnarray}
    $\{\mathcal{X}^{(q)}[l]=a_q(\mathcal{X}[l])\in \mathbb{R}^{M\times K \times N}, l\in \mathbb{Z}\}, \hspace{0.1in} q \in\{1,2\}$.
   % \end{eqnarray}
    We will illustrate particular choices of augmentations in the examples provided in Section \ref{sec:experiments}. We note that the proposed setting can be potentially extended to multi-modal schemes where the augmentations are defined for multiple modalities of the same input, such as visual and sound.
    \item We represent \underline{\it Siamese networks} to be trained by SSL with the mappings $f_q(\mathcal{U};\mathbb{W}^{(q)}_{DNN})$  for $q=1,2$, where $\mathcal{U}\in\mathbb{R}^{M\times K \times N}$ is the input, $\mathbb{W}^{(q)}_{DNN}$ %
    %=\{(\mathbf{W}_{DNN}^{(i)}, \mathbf{b}_{DNN}^{(i)}), i\in\{1,\ldots L_{DNN}\}\}$ 
    represents the trainable parameters of the $q^{\text{th}}$  network. 
  \item \underline{The outputs of the DNN} for different augmented inputs are represented by
   % \begin{eqnarray}
    $\mathbf{Y}^{(q)}[l]=f_q(\mathbb{X}^{(q)}[l];\mathbb{W}^{(q)}_{DNN})\in \mathbb{R}^{F\times N}, \hspace{0.1in} q\in\{1,2\}, l\in \mathbb{Z}$,
  %  \end{eqnarray}
    where $F$ is the output feature dimension.% 
    \item To be used in self-supervised training, we project the outputs of the enconders using \underline{\it projector networks}  $p_q(\mathbf{Y};\mathbb{W}^{(q)}_P)$, $ q=1,2$, where $\mathbf{Y}$ is its input, and $\mathbb{W}^{(q)}_P$ is the set of trainable parameters of the $q^{\text{th}}$ projector.
    \item \underline{The outputs of the projector network} for different augmentations are represented by
        %\begin{eqnarray}
    $\bZ^{(q)}[l]=p_q(\bY^{(q)}[l];\mathbb{W}^{(q)}_{P})\in \mathbb{R}^{P\times N}, \hspace{0.1in} q\in\{1,2\}, l\in \mathbb{Z}$, 
  %  \end{eqnarray}
    where $P$ is the projector head dimension.

\end{itemize}
For the numerical experiments in the article, we consider the weight-sharing setup where both encoders use the same network weights. We provide more details in Appendix \ref{sec:app_wsssl}. 

\subsection{Correlative mutual information maximization}
\label{sec:corInfoMax}
We define the CorInfoMax approach for SSL through the optimization problem
\begin{maxi!}[l]<b>
{\mathbb{W}_{DNN},\mathbb{W}_{P}}
{\hat{J}(\bz^{(1)};\bz^{(2)})[l]\label{eq:ldmiobjective}}{\label{eq:ldmioptimization}}{}
\end{maxi!}
where $\hat{J}^{(\varepsilon)}$ is the sample based estimate of (\ref{eq:stochcorinfomax})  at the $l^{\textrm{th}}$-batch which can be written as
\begin{eqnarray}
{\hat{J}(\bz^{(1)},\bz^{(2)})[l]=\log\det(\hat{\bR}_{\bz^{(1)}}[l]+\varepsilon \mathbf{I})+\log\det(\hat{\bR}_{\bz^{(2)}}[l]+\varepsilon\mathbf{I})-\frac{2}{\varepsilon N}\|\bZ^{(1)}[l]-\bZ^{(2)}[l]\|_F^2, \label{eq:covdet2}}
\end{eqnarray}
where the rightmost term stands for the sample based estimate of the term $2\varepsilon^{-1}E(\|\bz^{(1)}-\bz^{(2)}\|_2^2)$ in (\ref{eq:stochcorinfomax}), and $\hat{\bR}_{\bz^{(1)}}[l]$, $\hat{\bR}_{\bz^{(2)}}[l]$ and $\hat{\bR}_{\bz^{(1)}\bz^{(2)}}[l]$ are the auto-covariance matrix  and cross-covariance matrix estimates for the projector heads at the $l^{\textrm{th}}$ batch. If the batch size is large enough, these estimates can only be  obtained from the current batch samples. However, due to hardware limitations and considering the fact that intermediate batch sizes offer better accuracy \citep{mccandlish2018empirical,masters2018revisiting}, sufficiently-large batch sizes may not be possible to obtain reliable covariance estimates. Therefore, we adopt recursive covariance matrix estimation across batches \citep{rosca2006independent}. The corresponding covariance update expressions take the form
\begin{eqnarray}
\hat{\bR}_{\bz^{(q)}}[l]&=&\lambda \hat{\bR}_{\bz^{(q)}}[l-1]+(1-\lambda)\frac{1}{N}\tilde{\bZ}^{(q)}[l]\tilde{\bZ}^{(q)}[l]^T. \hspace{0.02in} q\in\{1,2\}, \label{eq:autocovupdate}
\end{eqnarray}
where $0\le\lambda<1$ is the forgetting factor, and $\tilde{\bZ}^{(q)}[l]$ represents the batch of mean-centralized projector outputs defined as 
%\begin{eqnarray}
$\tilde{\bZ}^{(q)}[l]={\bZ}^{(q)}[l]-{\bmu}^{(q)}[l]\mathbf{1}_N^T$,
%\end{eqnarray}
where $\{\bmu^{(q)}[l], q\in\{1,2\}\}$ represents the mean estimates for the projector outputs, updated by
%\begin{eqnarray}
$\bmu^{(q)}[l]=\lambda \bmu^{(q)}[l-1]+(1-\lambda)\frac{1}{N}\bZ^{(q)}{\mathbf{1}_N}$, for  $q\in\{1,2\}$.
%\end{eqnarray}
%Similar to the auto-covariance matrix update in (\ref{eq:autocovupdate}),
We can write the update expression for the cross-covariance matrix as
%\begin{eqnarray}
$\hat{\bR}_{\bz^{(1)}\bz^{(2)}}[l]=\lambda \hat{\bR}_{\bz^{(1)}\bz^{(2)}}[l-1]+(1-\lambda)\frac{1}{N}\tilde{\bZ}^{(1)}[l]\tilde{\bZ}^{(2)}[l]^T$. (See Appendix \ref{app:pseudocode} for CorInfoMax pseudocode.) %\nonumber %\label{eq:crosscovupdate}
%\end{eqnarray}

It is informative to inspect the terms in the objective function $\hat{J}(\bz^{(1)},\bz^{(2)})$ in (\ref{eq:covdet2}):
\begin{itemize}
    \item Minimization of $\frac{2\varepsilon^{-1}}{N}\|\bZ^{(1)}[l]-\bZ^{(2)}[l]\|_F^2$, acts as a force to pull the representations of alternative augmentations toward each other, which we refer to as the {\it attraction factor}. %Figure \ref{fig:toypicture}.(c) illustrates the gradients corresponding to this cost function (represented by white arrows).
    \item Maximization of $\log\det(\hat{\bR}_{\bz^{(1)}}[l]+\varepsilon\mathbf{I})$ acts as a dispersion force causing non-degenerate (for $\varepsilon \approx 0$) expansion of projection vectors in the $P$-dimensional space, which we informally refer to as  {\it big-bang factor}. 
    Therefore, the covariance determinant acts as a regularization or barrier function avoiding collapse in the latent space. Consequently, it provides a convincing replacement for negative samples in contrastive methods to prevent feature collapse. Appendix \ref{app:eigs} illustrates the eigenvalues of the projector covariance vector for the CIFAR-10 training, which demonstrates that the CorInfoMax criterion spreads information throughout the feature space, avoiding dimensional collapse. 
\end{itemize}
 \looseness=-1 Figure \ref{fig:toypicture}.(c) illustrates the learning dynamics of the CorInfoMax approach based on (\ref{eq:covdet2}) with a toy picture. In this figure, the ellipsoid is a representative surface for the level set of the quadratic function $(\bz-{\bmu}^{(q)})^T\mathbf{R}_{\mathbf{z}^{(q)}}^{-1}(\bz-{\bmu}^{(q)})$, which reflects the spreading pattern of the latent vectors around their mean. The black arrows represent the gradient of the $\log\det(\mathbf{R}_{\mathbf{z}^{(q)}})$ regularization factor, which acts as a force to push latent vectors away from the center of the ellipsoid ($\bmu^{(q)}$) and therefore corresponds to the expansion force of the "big-bang" factor. In the same figure, the white arrows correspond to the gradients corresponding to the attraction factor, i.e., the Euclidian distance between two positive pairs. In summary, we can view the learning dynamics of CorInfomax SSL as an expansion in the latent space, while the representations of positive samples are attracted to each other.
More information on the CorrInfoMax optimization setting is provided in the Appendix \ref{app:suppcorinfomax}. Furthermore, Appendix \ref{app:embeding} contains a visualization of the embeddings obtained using the CorInfoMax criterion.

\subsection{Computational complexity}
\label{sec:complexity}
The main difference between our and related SSL methods is the extra log-determinant computation. The log determinant is typically computed using a decomposition method such as the LU decomposition, whose gradient requires a matrix inversion. Both  determinant and  inversion have the same complexity as matrix multiplication \citep{bunch1974triangular}, i.e. $O(n^\alpha)$ with $2 < \alpha\leq 3$ for multiplying two $n\times n$ matrices. In our experiments with GPUs, we observe an almost flat runtime cost up to $n=1024$, which implies that the overhead due to the $\log\det$ cost is insignificant in practice.   We find the extra cost of log determinant and its gradient to be negligible compared to the rest of the model computation, whose runtime is dominated by the encoder, as confirmed by the runtime experiments reported in Appendix \ref{app:runtime}.

%% file: blockdiagram2branchv2.tex
\begin{center}
%\begin{scaletikzpicturetowidth}{0.7\textwidth}
%\resizebox{\textwidth}{!}{%
{
\begin{tikzpicture}[thick,scale=0.7, every node/.style={transform shape}]
%\begin{tikzpicture}[thick,scale=0.8, every node/.style={scale=0.8}]
  % The order of blocks matters since some are partially hidden behind subsequent blocks.
  %%%%%%%%%%%%%% INPUT BLOCK %%%%%%%%%%%%%%%%%%%%%%
  \node[inputbnew,text width=1cm](input1){{\large Input \\ \vspace{0.1in}\hspace{0.03in} {\Large $\mathcal{X}$}}};
  \Dimline[($(input1)+(-0.25,2)$)][($(input1)+(1.25,2)$)][{\large N}];
  \DimlineN[($(input1)+(-1.05,1.2)$)][($(input1)+(-1.05,-1.2)$)][{\large M}];
  \DimlineKnew[($(input1)+(0.85,-1.3)$)][($(input1)+(1.35,-0.8)$)][{ K}];
  %\node[block,right=1cm of input1,yshift=-2.5cm](augm1){{Augmentation 2}};
  %%%%%%%%%%%% AUGMENTATION 1 %%%%%%%%%%%%%%%%%%%%%%
  \node[draw,
    fill=coldblue,
    minimum width=2cm,
    minimum height=1.2cm,
    right=1.5cm of input1,
    yshift=75.5, 
    rounded corners](augm1) {\color{black}\LARGE $a_1(\cdot)$};
    \node[right=-2.3cm of augm1, yshift=25]{Augmentation-1};
  %%%%%%%%%%%% AUGMENTATION 2 %%%%%%%%%%%%%%%%%%%%%%
  \node[draw,
    fill=coldblue,
    minimum width=2cm,
    minimum height=1.2cm,
    right=1.5cm of input1,
    rounded corners,
    yshift=-65.5](augm2) {\color{black}\LARGE $a_2(\cdot)$};
    \node[right=-2.3cm of augm2, yshift=25]{Augmentation-2};
    %%%%%%%%%%%% INPUT TO AUGMENTATION 1 %%%%%%%%%%%%%%%%%%%%%%
    \node[right=0.1cm of input1](l1s){};
    \node[right=0.95cm of input1](l1e){};
    \draw[-,darkgray,thick] (l1s.center) -- (l1e.center);
    \node[right=0.95cm of input1,yshift=75.5](l2e){};
    \node[right=0.1cm of l2e](l3e){};
    \draw[-,darkgray,thick] (l1e.center) -- (l2e.center);
    \draw[->,darkgray,thick] (l2e.center) -- (augm1.west);
    %%%%%%%%%%%% INP TO AUGMENTATION 2 %%%%%%%%%%%%%%%%%%%%%%
    \node[right=0.95cm of input1,yshift=-65.5](l4e){};
    \draw[-,darkgray,thick] (l1e.center) -- (l4e.center);
    \node[right=0.1cm of l4e](l5e){};
    \draw[->,darkgray,thick] (l4e.center) -- (augm2.west);
    %%%%%%%%%%%% AUGMENTATION 1 OUTPUT %%%%%%%%%%%%%%%%%%%%%%
    \node[inputbAnew,text width=1cm,right=0.5cm of augm1,yshift=0](inputa1){{{\Large $\mathcal{X}^{(1)}$}}};
    %%%%%%%%%%%% AUGMENTATION 2 OUTPUT %%%%%%%%%%%%%%%%%%%%%%
    \node[inputbBnew,text width=1cm,right=0.5cm of augm2,yshift=0](inputa2){{{\Large $\mathcal{X}^{(2)}$}}};
    % Augmentation 1 output lines
    \node[right=1.75cm of l3e](l6b){};
    \node[right=2.8cm of l3e](l6e){};
    \node[left=0.45cm of inputa1.center](nl6e){};
    %\draw[->,red,thick] (l6b.center) -- (l6e.center);
    \draw[->,darkgray,thick] (augm1.east) -- (nl6e);
    \node[left=0.45cm of inputa2.center](nl7e){};
    \draw[->,darkgray,thick] (augm2.east) -- (nl7e);
    %\node[below=0.95cm of l6e](l7e){};
    %\draw[->,red,thick] (l6e.center) -- (l7e.center);
    % Augmentation 2 output lines
   % \node[right=1.75cm of l5e](l8b){};
%    \node[right=3.8cm of l5e](l8e){};
 %   \draw[-,red,thick] (l8b.center) -- (l8e.center);
  %  \node[above=0.85cm of l8e](l9e){};
  %  \draw[->,red,thick] (l8e.center) -- (l9e.center);
  \node[draw,
    fill=coldblue,
    minimum width=2cm,
    minimum height=1.2cm,
    right=1cm of inputa1,
    yshift=2,
    rounded corners](dnn1) {\color{black}\Large $f_1(\cdot; \mathbb{W}^{(1)}_{DNN})$};
    \node[draw,
    fill=coldblue,
    minimum width=2cm,
    minimum height=1.2cm,
    right=1cm of inputa2,
    yshift=2,
    rounded corners](dnn) {\color{black}\Large $f_2(\cdot; \mathbb{W}^{(2)}_{DNN})$};
    \node[right=-2.0cm of dnn1, yshift=25]{DNN-1};
    \node[right=0.1cm of inputa1,yshift=2](nl10s){};
    \node[right=0.9cm of inputa1,yshift=2](nl10e){};
    \draw[->,darkgray,thick] (nl10s.center) -- (nl10e.center);
    \node[right=-2.0cm of dnn, yshift=25]{DNN-2};
    \node[right=0.1cm of inputa2,yshift=2](l10s){};
    \node[right=0.9cm of inputa2,yshift=2](l10e){};
    \draw[->,darkgray,thick] (l10s.center) -- (l10e.center);
    \node[draw,
    fill=coldblue,
    minimum width=2cm,
    minimum height=1.2cm,
    right=0.5cm of dnn1,
    yshift=0,
    rounded corners](proj1) {\color{black}\Large $p_1(\cdot; \mathbb{W}^{(1)}_{P})$};
    \node[draw,
    fill=coldblue,
    minimum width=2cm,
    minimum height=1.2cm,
    right=0.5cm of dnn,
    yshift=0,
    rounded corners](proj) {\color{black}\Large $p_2(\cdot; \mathbb{W}^{(2)}_{P})$};
    \node[right=-2.1cm of proj1, yshift=25]{Projector-1};
    \node[right=-2.1cm of proj, yshift=25]{Projector-2};
    \draw[->,darkgray,thick] (dnn1.east) -- (proj1.west);
    \draw[->,darkgray,thick] (dnn.east) -- (proj.west);
    \node[draw,
    top color=coldgreen2,
    bottom color=coldgreen2,
    minimum width=1.5cm,
    minimum height=2cm,
    right=0.5cm of proj1](pout1){\color{black}\large$\mathbf{Z}^{(1)}$};
    \node[draw,
    top color=coldorange,
    bottom color=coldorange,
    minimum width=1.5cm,
    minimum height=2cm,
    right=0.5cm of proj](pout2){\color{black}\large$\mathbf{Z}^{(2)}$};
    \draw[->,darkgray,thick] (proj1.east) -- (pout1.west);
    \draw[->,darkgray,thick] (proj.east) -- (pout2.west);
    %\node[draw,
    %top color=bluegray,
    %bottom color=red,
    %minimum width=1.5cm,
    %minimum height=2cm,
    %right=0.0cm of pout1](pout2){\color{white}\large$\mathbf{Z}^{(2)}$};
    \DimlineP[($(pout1)+(1.05,1.05)$)][($(pout1)+(1.05,-1.05)$)][{\large P}];
    \DimlineP[($(pout2)+(1.05,1.05)$)][($(pout2)+(1.05,-1.05)$)][{\large P}];
   \Dimline[($(pout1)+(-0.69,1.15)$)][($(pout1)+(0.75,1.15)$)][{\large N}];
    \Dimline[($(pout2)+(-0.69,1.15)$)][($(pout2)+(0.75,1.15)$)][{\large N}];
    % Classifier
  %  \node[right=0.11cm of dnn,yshift=3](c1s){};
   % \node[right=-0.25cm of c1s,yshift=-70](c1e){};
%    \draw[dotted,red,thick] (c1s) -- (c1e);
 %   \node[draw,
 %   fill=junglegreen,
  %  minimum width=2cm,
   % minimum height=1.2cm,
    %right=0.3cm of c1e,
    %yshift=0](class) {\color{white}\Large $c(\cdot; \mathbb{W}_{C})$};
    %\draw[dotted,->,red,thick] (c1e.center) -- (class.west);
    %\node[right=-2.3cm of class, yshift=-25]{Linear Classifier};
\end{tikzpicture}
}
%\end{scaletikzpicturetowidth}
\end{center}

%% file: experimentsreb2.tex
\subsection{Implementation details} \label{sssec:impl}

\looseness=-1\underline{\it Datasets:} We perform experiments on CIFAR-10, CIFAR-100 \citep{krizhevsky2009learning}, Tiny ImageNet \citep{Le2015TinyIV}, COCO \citep{lin2014coco}, ImageNet-100 and ImageNet-1K \citep{5206848} datasets\footnote{CorInfoMax's source code is publicly available in \url{https://github.com/serdarozsoy/corinfomax-ssl}}. For ImageNet-100, we use the same subset of ImageNet-1K \citep{5206848} as related work \citep{tian2019contrastive,kalantidis2020hard,ge2021robust}. (See Appendix \ref{app:datasets} for more details.)

\underline{\it Training Procedure: } The experiments consist of two consecutive stages: pretraining and linear evaluation. We first perform the unsupervised pretraining  of the encoder network $f$ by applying the proposed CorInfoMax method described in Section \ref{sec:corInfoMax} on the training dataset. After completing  pretraining, we perform the linear evaluation, a standardized protocol to evaluate the quality of the learned representations \citep{kolesnikov2019revisiting, chen2020simple, grill2020bootstrap}. 

For the linear evaluation stage, we first perform supervised training of the linear classifier using the representations obtained from the encoder network $f$ with frozen coefficients on the same training dataset. Then, we obtain the test accuracy results for the trained linear classifier based on the validation dataset. 

\underline{\it Computational Resources:} For ImageNet-100 and ImageNet-1K, we pretrain our model on up to $8$ A100 Cloud GPUs. The remaining datasets are trained using a single T4 and V100 Cloud GPU. Linear evaluations are performed using the same type and amount of computational resources. The details related to batch sizes are described in \ref{subsec:optim}.
%Subsections can be deleted and all the things can be collected in one part

\subsubsection{Input augmentations}

During the \underline{\it pretraining stage}, two augmented versions of each input image are generated as shown in Figure \ref{fig:ssblockdiagram}.
During this process, each image is cropped with random size, resized to the original resolution, followed by random applications of horizontal mirroring, color jittering, grayscale conversion, Gaussian blurring, and solarization. 
Since we use the same augmentation parameters as BYOL 
\citep{grill2020bootstrap} and VicReg \citep{bardes2021vicreg}: each augmentation branch uses the same probability values for these randomized operations except Gaussian blurring and solarization, which use different probabilities.

During the training phase of the \underline{\it linear evaluation} stage, a single augmentation of each input image is produced by random cropping and resizing followed by a random horizontal flip. For the test phase of linear evaluation, we use resize and center crop augmentations, similar to \citep{zbontar2021barlow,bardes2021vicreg}.
We provide more details about the augmentations in Appendix \ref{app:augmentation}.

\subsubsection{Network architecture}

\underline{\it Encoder Network}: For CIFAR datasets, we use a modified form of ResNet-18 architecture \citep{he2016deep} similar to \citep{chen2020simple, chen2021exploring, haochen2021provable}.We use standard ResNet-50(\citep{he2016deep}) for Tiny ImageNet, ImageNet-100 and ImageNet-1K, and also standard ResNet-18(\citep{he2016deep}) for ImageNet-100. In all cases, the last fully connected layer is removed. Therefore, the encoder output size is $512$ and $2048$ for ResNet-18 and ResNet-50, respectively. The encoder network $f$ shares weights between augmented branches (see Appendix \ref{sec:app_wsssl}).

\underline{\it Projector Network:} The output of the encoder network is fed into the projector network, as in Figure \ref{fig:ssblockdiagram}. The projector network $p$ is a $3$-layer MLP, with ReLU activation functions for the hidden layers and linear activation functions for the output layer. The projector dimensions are  $2048$-$2048$-$64$  for the CIFAR-10 dataset, $4096$-$4096$-$128$ for the CIFAR-$100$, Tiny ImageNet and ImageNet-$100$, and $8192$-$8192$-$512$ for the ImageNet-$1$K.
Finally, we perform the $L_{2}$-normalization on the projector output.

\underline{\it Linear Classifier:} For the linear evaluation phase, we employ a standard linear classifier whose input is the weight-frozen encoder network's output.

\subsubsection{Optimization} \label{subsec:optim}

For pretraining, we use $1000$ epochs with a batch size of $512$ for CIFAR, and $800$ epochs with a batch size of $1024$ for Tiny ImageNet. For ImageNet-100 experiments, we use 400 epochs for ResNet-18 and 200 epochs for ResNet-50, with a batch size of 1024 for both. ImageNet-1K experiments are conducted as 100 epochs with a batch size of 1536. We use the SGD optimizer with a momentum of $0.9$ and a weight decay of $1e-4$. The initial learning rate is $0.5$ for CIFAR datasets and Tiny ImageNet, $1.0$ for ImageNet-100, and $0.2$ for ImageNet-1K. These learning rates follow the cosine decay with a linear warmup schedule.

We use the modified form of (\ref{eq:covdet2}) as our objective function, where we replace $\frac{2\varepsilon^{-1}}{N}$ with $\alpha$, the attraction coefficient, which we consider as a separate hyper-parameter. \looseness=-1 In our experiments, the diagonal perturbation is $\varepsilon=1\mathrm{e}{-8}$, while $\alpha=250$ for CIFAR-10, $\alpha=1000$ for CIFAR-100, $\alpha=2000$ for ImageNet-1K, and $\alpha=500$ for Tiny ImageNet and ImageNet-100. The forgetting factor $\lambda=0.01$ for all datasets except Tiny ImageNet and ImageNet-1K, which have $\lambda=0.1$. Details with coefficients of our loss function are provided in Appendix \ref{app:parameters}.

\looseness=-1 For  linear evaluation, the linear classifier is trained for $100$ epochs with a batch size of $256$ for all datasets. We used the SGD optimizer with a momentum of $0.9$, and without weight decay.  For all datasets except ImageNet-$1$K, cosine decay schedule is utilized with initial and minimum learning rates of $0.2$ and $2\mathrm{e}{-3}$ respectively. For Imagenet-$1$K, we use a step scheduler with a starting value of $25$, which is reduced by a factor of $10$ every $20$ epoch.

\subsection{Results}

We evaluate the learned representations from the CorInfoMax pretraining by following the linear evaluation protocol explained in Sec. \ref{sssec:impl}. Table \ref{evaluation-table-1} shows that CorInfoMax achieves state-of-the-art performance in linear classification after pretraining. Due to the limited size of the validation sets (5-10K), differences less than $\approx 0.5\%$ are not statistically significant. The full comparison is provided in Table \ref{extended-evaluation-table-1} in Appendix \ref{app:fullcomparison}. It is also interesting to observe the progress of the LDMI measure during the CorInfoMax training process, which is illustrated in Appendix \ref{app:minfoevolution}.

\begin{table}[h]
  \caption{Top-1 accuracies {($\%$)} under linear evaluation on different datasets. Results are reported from \citep{da2022solo, ermolov2021whitening,haochen2021provable} for CIFAR-10 and CIFAR-100, \citep{haochen2021provable} for Tiny ImageNet,\citep{ge2021robust, lee2021improving} for ImageNet-100 (IN-100) in in ResNet-50,  \citep{da2022solo} for ImageNet-100 (IN-100) in ResNet-18, \citep{haochen2021provable, chen2021exploring} for ImageNet-1K (IN-1K). In the case of the result of a model in more than one resource, we integrate the largest score. We \textbf{bold} all top results  that are statistically indistinguishable.} 
  \label{evaluation-table-1}
  \centering
  \begin{tabular}{@{}l@{\hspace*{1mm}}cccc@{\hspace*{2mm}}ccc@{}}
    \toprule
    \multirow{2}[2]{*}{Method} & \multicolumn{1}{c}{CIFAR-10} &  \multicolumn{1}{c}{CIFAR-100} & \multicolumn{1}{c}{Tiny-IN}& \multicolumn{2}{c}{IN-100} & \multicolumn{2}{c}{IN-1K} \\
    \cmidrule(l){2-2} \cmidrule(l){3-3} \cmidrule(l){4-4}  \cmidrule(l){5-6} \cmidrule(l){7-7}
    & {ResNet-18} & {ResNet-18} &  {ResNet-50} &  {ResNet-18} &  {ResNet-50} & {ResNet-50} &\\
    \midrule 
    SimCLR \citep{chen2020simple} & 91.80 & 66.83 &  48.12  & 77.04 & - & 66.5 &\\
    SimSiam \citep{chen2021exploring}& 91.40 & 66.04 & 46.76 & 78.72 & 81.6 & 68.1 &\\
    Spectral \citep{haochen2021provable} & 92.07 & 66.18 & 49.86  & - & - & 66.97 &\\
    BYOL \citep{grill2020bootstrap} & 92.58 &  70.46 & - & \textbf{80.32} & 78.76 & \textbf{69.3} &\\
    W-MSE 2 \citep{ermolov2021whitening} & 91.55 & 66.10 &  - & 69.06 & - & - &\\
    MoCo-V2 \citep{chen2020improved} & \textbf{92.94} & 69.89 &  - & 79.28 & - &  67.4 &\\
    Barlow \citep{zbontar2021barlow} & 92.10 & 70.90 & - & \textbf{80.38} & - & 68.7 &\\
    VICReg \citep{bardes2021vicreg} & 92.07 & 68.54 & - & 79.40 & - & 68.6 &\\
    \midrule 
    CorInfoMax & \textbf{93.18} & \textbf{71.61}  &  \textbf{54.86} &  \textbf{80.48} &  \textbf{82.64} & {69.08} &\\
    \bottomrule
  \end{tabular}
\end{table}

We compare the semi-supervised learning performance of CorInfoMax with VICReg \citep{bardes2021vicreg}. For a fair comparison, VICReg is pretrained for 100 epochs on ImageNet-1K, then semi-supervised learning performances of both models are evaluated by fine-tuning the encoders with $1(\%)$ and $10(\%)$ of the labeled ImageNet-1K dataset. The results are presented in Table \ref{semi-table-1}, while details are provided in Appendix \ref{app:imagenet1k-hyp}.

\begin{table}[h]
  \caption{ Top-1 accuracies {($\%$)} under semi-supervised classification on ImageNet-1K dataset after 100 epoch pretraining. VICReg is pretrained and evaluated using hyper-parameters reported in \citep{bardes2021vicreg}.}
  \label{semi-table-1}
  \centering
  \begin{tabular}{lccccc}
    \toprule
    Method &  {1$\%$ of samples} & {10$\%$ of samples} \\
    \midrule 
    {VICReg} & 44.75 & 62.16 \\
    {CorInfoMax} & 44.89 & 64.36 \\
    \bottomrule
  \end{tabular}
\end{table}

We experimented with transfer learning on object detection and instance segmentation with the same procedure and code of MoCo\citep{he2019moco}, and reproduce the results with the MoCo V2 \citep{chen2020mocov2} checkpoint. Our algorithm shows competitive performance with MoCo V2 \citep{chen2020mocov2}, related details are provided in Appendix \ref{app:transfer}.

%% file: appendixreb2.tex
\renewcommand{\theequation}{A.\arabic{equation}}
\setcounter{equation}{0}
\renewcommand{\thefigure}{A.\arabic{figure}}
\setcounter{figure}{0}
\section{Background on information measures}
\label{app:infomeas}
The most natural and conventional means to measure uncertainty of a real-valued random vector $\bx$ is to use the joint Shannon differential entropy defined by \citep{thomas2006elements}
\begin{eqnarray}
h(\bx)&=&-\int\displaylimits_{\text{dom}(f_\bx)} \log(f_\bx(x))f_\bx(x)\mathrm{d}x= -E(\log(f_\bx(\bx)))\nonumber,
\end{eqnarray}
where $f_\bx$ is the joint probability density function (pdf) of the components of $\bx$. The corresponding Shannon mutual information (SMI) between the random vectors $\bx$ and $\by$ is defined as
\begin{eqnarray}
I(\bx;\by)&=& h(\bx)-h(\bx \cvert \by) \label{eq:smi}
\end{eqnarray}
where $h(\bx \cvert \by)=-E(\log(f_{\bx \cvert \by}(\bx \cvert \by)))$ is the conditional entropy. Shannon mutual information is a measure of the dependence between its arguments.

For a given $r$-dimensional random vector $\bx$ with the pdf $f_\bx(\bx)$, log-determinant (LD) entropy is defined as \citep{zhouyin:21,erdogan2022blind}
\begin{eqnarray}
\label{eq:HLD}
\hld(\bx)=\frac{1}{2}\log\det(\bR_\bx+\varepsilon \bI)+\frac{r}{2}\log(2\pi e),
\end{eqnarray}
where $\bR_\bx$ is the auto-covariance matrix of $\bx$, and $\varepsilon$ is a small nonnegative parameter 
defined for the diagonal perturbation on $\bR_\bx$.  We note that $\hld(\bx)$ is equivalent to Shannon's entropy when $\bx$ is a Gaussian vector with covariance $\bR_\bx$ and $\varepsilon$ is equal to zero. However, we should underline that it is a standalone uncertainty measure solely based on the second-order statistics, which reflects linear dependence.

The joint LD-entropy of an $r$-dimensional random vector $\bx$ and a $q$-dimensional random vector $\by$ is defined as the LD-entropy of the cascaded vector $\left[\begin{array}{cc} \bx^T & \by^T \end{array}\right]^T$, i.e., 
\begin{eqnarray*}
\hld(\left[\begin{array}{cc} \bx^T & \by^T \end{array}\right]^T)=\frac{1}{2}\log\det\left(\bR_{\left[\begin{array}{c} \bx\\  \by \end{array}\right]}+\varepsilon \bI\right)+\frac{r+q}{2}\log(2\pi e)
\end{eqnarray*}
where 
$$
\bR_{\left[\begin{array}{c} \bx\\  \by \end{array}\right]}=\left[\begin{array}{cc} \bR_\bx & \bR_{\bx\by}\\ \bR_{\bx\by}^T & \bR_\by\end{array}\right]
$$
is the covariance matrix of $\left[\begin{array}{cc} \bx^T & \by^T \end{array}\right]^T$, $\bR_\by$ is the auto-covariance matrix of $\by$, and $\bR_{\bx\by}$ is the cross-covariance matrix of $\bx$ and $\by$. Using the determinant decomposition based on Schur's complement \cite{kailath2000linear}, we can write
\begin{eqnarray}
\hld(\left[\begin{array}{cc} \bx^T & \by^T \end{array}\right]^T)&&=\frac{1}{2}\log\det(\bR_\by+\varepsilon \bI)+\frac{q}{2}\log(2\pi e)\nonumber \\
&&+\frac{1}{2} \log\det(\bR_\bx-\bR_{\bx\by}(\bR_\by+\varepsilon\mathbf{I})^{-1}\bR_{\bx\by}^T+\varepsilon\mathbf{I})+\frac{r}{2}\log(2\pi e) \nonumber \\
&&=\hld(\by)+\hld(\bx|\by), \nonumber
\end{eqnarray}
where we defined 
\begin{eqnarray}
\hld(\bx|\by)=\frac{1}{2}\log\det(\bR_\bx-\bR_{\bx\by}(\bR_\by+\varepsilon\mathbf{I})^{-1}\bR_{\bx\by}^T+\varepsilon\mathbf{I})+\frac{r}{2}\log(2\pi e), \label{eq:hldcond}
\end{eqnarray}
as the conditional LD-entropy.
 We note that the argument of the $\log\det$ function in (\ref{eq:hldcond}) is the auto-covariance of the error for linearly estimating $\bx$ from $\by$ with respect to the minimum mean square estimation (MMSE) criterion for $\varepsilon \to 0$ (see, for example, Theorem 3.2.2 of \cite{kailath2000linear}). More precisely, given that $\mu_\bx$ and $\mu_\by$ represent the means of $\bx$ and $\by$, respectively,  $\hat{\bx}_{MMSE}=\bR_{\bx\by}\bR_{\by}^{-1}(\by-\mu_\by)+\mu_\bx$ is the best linear MMSE estimate of $\bx$ from $\by$. Therefore, if we define $\be_{MMSE}=\bx-\hat{\bx}_{MMSE}$, the auto-covariance matrix of $\be_{MMSE}$ is given by $\bR_{\be_{MMSE}}=\bR_\bx-\bR_{\bx\by}\bR_{\by}^{-1}\bR_{\bx\by}^T$, which is the argument of the $\log\det$ function in (\ref{eq:hldcond}) for $\varepsilon \to 0$.  As a result, we can view $\hld(\bx|\by)$, which is the LD-Entropy of $\be_{MMSE}$, as a measure of the remaining uncertainty after linearly (affinely) estimating $\bx$ from $\by$ based on the MMSE criterion. 

The LD-mutual information (LDMI) measure is defined based on (\ref{eq:HLD}) and (\ref{eq:hldcond}) as follows:
\begin{eqnarray}
\ild(\bx;\by)&=&\hld(\bx)-\hld(\bx \cvert \by) \nonumber \\
  &=& \frac{1}{2}\log\det(\bR_\bx+\varepsilon \mathbf{I})-\frac{1}{2}\log\det(\bR_\bx-\bR_{\bx\by}(\bR_\by+\varepsilon\mathbf{I})^{-1}\bR_{\bx\by}^T+\varepsilon\mathbf{I}). \label{eq:ldmi}
\end{eqnarray}
Taking the average of (\ref{eq:ldmi}) with its symmetric version obtained by exchanging $\bx$ and $\by$, we can obtain an alternative but equivalent expression for LDMI:
\begin{eqnarray}
\ild(\bx;\by)&=&\frac{1}{4}\log\det(\bR_\bx+\varepsilon \mathbf{I})+\frac{1}{4}\log\det(\bR_\by+\varepsilon \mathbf{I}) \nonumber \\&&-\frac{1}{4}\log\det(\bR_\bx-\bR_{\bx\by}(\bR_\by+\varepsilon\mathbf{I})^{-1}\bR_{\bx\by}^T+\varepsilon\mathbf{I})\nonumber \\
&& -\frac{1}{4}\log\det(\bR_\by-\bR_{\by\bx}(\bR_\bx+\varepsilon\mathbf{I})^{-1}\bR_{\by\bx}^T+\varepsilon\mathbf{I}). \label{eq:ldmi3a}
\end{eqnarray}

The following lemma asserts that $\ild(\bx,\by)$ is an information measure reflecting the correlation or ``linear dependence" between two vectors \citep{erdogan2022blind}:

\begin{lemma}
Let $\bx,\by$ be random vectors with the auto-covariance matrices $\bR_\bx>0$ and $\bR_\by$, respectively, and the cross-covariance matrix $\bR_{\bx\by}$. Then,
\begin{itemize}
    \item $\ild(\bx;\by)\ge 0$,
    \item $\ild(\bx;\by)=0$ if and only if $\bR_{\bx\by}=\mathbf{0}$, that is, $\bx$ and $\by$ are uncorrelated.
\end{itemize}
\end{lemma}

\section{The weight-shared SSL setup}
\label{sec:app_wsssl}
\begin{figure}[h]
\hspace*{-0.2in}\input{blockdiagram.tex}
\caption{Self-supervised learning set-up with weight sharing.}
\label{fig:ssblockdiagram2}
\end{figure}
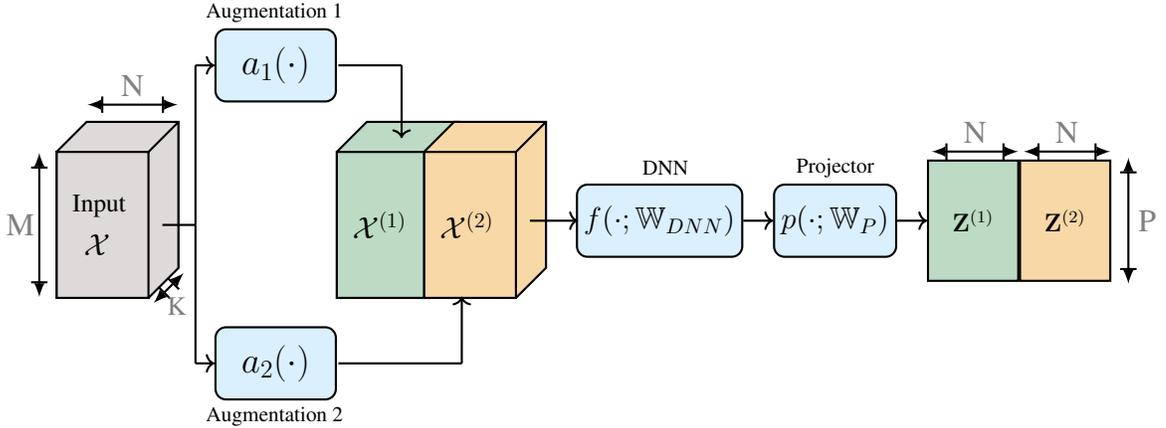

In all our experiments we use weight-sharing between two branches: 
\begin{eqnarray*}
%\mathbb{W}_{DNN}^{(1)}&=&\mathbb{W}_{DNN}^{(2)}=\mathbb{W}_{DNN}\\
%\mathbb{W}_{P}^{(1)}&=&\mathbb{W}_{P}^{(2)}=\mathbb{P}_{DNN}\\
f_1(\cdot; \mathbb{W}_{DNN}^{(1)})&=&f_2(\cdot; \mathbb{W}_{DNN}^{(2)})=f(\cdot; \mathbb{W}_{DNN})\\
p_1(\cdot; \mathbb{W}_{P}^{(1)})&=&p_2(\cdot; \mathbb{W}_{P}^{(2)})=p(\cdot; \mathbb{W}_{P}),
\end{eqnarray*}
The two-branch network in Figure \ref{fig:ssblockdiagram} is equivalent to the setup illustrated in Figure \ref{fig:ssblockdiagram2}. The augmentation outputs $\mathcal{X}^{(1)}$ and $\mathcal{X}^{(2)}$ are input to the same DNN, i.e., $f(\cdot; \mathbb{W}_{DNN})$ whose output is input to the projection network $p(\cdot ; \mathbb{W}_{P})$. 

\section{Pseudocode}
\label{app:pseudocode}
Algorithm \ref{algo:corinfomax} (next page) describes the main steps for the CorInfoMax approach in a PyTorch-style pseudocode format.

\begin{algorithm}[h]
\SetAlgoLined
    \PyComment{f: encoder with projector network} \\
    \PyComment{N: batch size, D: projector output dimension} \\
    \PyComment{R1 and R2: covariance matrices are initialized as identity (DXD)} \\
    \PyComment{mu1 and mu2: means vectors are initiliazed as zero (D)} \\
    \PyComment{lambda: forgetting factor, alpha: attraction coefficient} \\
    \PyComment{mse\_loss: mean squared error loss, @: matrix multiplication} \\
    \PyCode{for x in loader:} \PyComment{load input batch} \\ 
    \Indp   % start indent
        \PyComment{random augmentations} \\
        \PyCode{x1, x2 = augmentation(x)} \\ 
        \PyComment{projector outputs}\\
        \PyCode{z1 = f(x1)}  \\ 
        \PyCode{z2 = f(x2)}  \\ 
        \PyComment{mean estimation} \\
        \PyCode{mu1\_update = z1.mean(0)}  \\ 
        \PyCode{mu2\_update = z2.mean(0)}  \\ 
        \PyCode{mu1 = lambda * mu1 + (1 - lambda) * mu1\_update} \\ 
        \PyCode{mu2 = lambda * mu2 + (1 - lambda) * mu2\_update}  \\ 
        \PyComment{covariance matrix estimation} \\
        \PyCode{z1\_hat = z1 - mu1}  \\ 
        \PyCode{z2\_hat = z2 - mu2}  \\ 
        \PyCode{R1\_update = (z1\_hat.T @ z1\_hat) / N}  \\ 
        \PyCode{R2\_update = (z2\_hat.T @ z2\_hat) / N}  \\ 
        \PyCode{R1 = lambda * R1 + (1 - lambda) * R1\_update}  \\ 
        \PyCode{R2 = lambda * R2 + (1 - lambda) * R2\_update}  \\ 
        \PyComment{loss calculation} \\
        \PyCode{cor\_loss = - (logdet(R1) + logdet(R2)) / D}  \PyComment{bing-bang factor}\\  
        \PyCode{sim\_loss = mse\_loss(z1, z2)} \PyComment{attraction factor}\\ 
        \PyCode{loss = cor\_loss + alpha * sim\_loss} \\
        \PyComment{optimization} \\
        \PyCode{loss.backward()} \\
        \PyCode{optimizer.step()} \\
    \Indm 
\caption{PyTorch-style pseudocode for CorInfoMax}
\label{algo:corinfomax}
\end{algorithm}

\section{Datasets}
\label{app:datasets}
% As dataset properties,
\begin{itemize}
\item The \textit{CIFAR-$10$} dataset \citep{krizhevsky2009learning} consists of $32\times32$ images with  $10$ classes. There are $5000$ training images and $1000$ validation images for each class.
\item The \textit{CIFAR-$100$} dataset \citep{krizhevsky2009learning} consists of $32\times32$ images with  $100$ classes. There are $500$ training images and $100$ validation images for each class.
\item The \textit{Tiny ImageNet} dataset \citep{Le2015TinyIV} consists of $64\times64$ images with  $200$ classes. There are $500$ training images and $50$ validation images for each class.
\item \textit{ImageNet-$1$K} \citep{5206848} has 1281167 different sizes of images from 1000 classes as training set. Set of 50000 validation images is treated as a test dataset for evaluation purposes. 
\item The \textit{ImageNet-$100$} dataset \citep{tian2019contrastive,kalantidis2020hard, ge2021robust}  contains $100$ sub-classes of the ImageNet dataset \citep{5206848}, which consists of  images with variety of sizes. There are $1300$ train images and $50$ validation images for each class.
\item The \textit{COCO} dataset \citep{lin2014coco} consists of 164K images with annotations for object detection, instance segmentation, captioning, keypoint detection, and per-pixel segmentation. The training split contains 118K images, while the validation set contains 5K images. The test set contains 41K images. 

\end{itemize}

Images in all these datasets have three color channels. 

\section{Image augmentations}
\label{app:augmentation}

\subsection{Augmentations during pretraining}
During CorInfoMax pretraining, we use the following set of augmentations:
\begin{itemize}
  \item Random resized cropping: cropping a random area of the input image with a scale parameter  $(0.08,1.0)$. Then resizing that cropped area to $32\times32$ for CIFAR datasets, $64\times64$ for Tiny ImageNet, and $224\times224$ for ImageNet-100 and ImageNet-1K. 
  \item Horizontal flipping: mirroring the input image horizontally (left-right).
  \item Color jittering: changing the color properties of the input image. Brightness, contrast, and saturation are (uniform) randomly selected from $[max(0, 1 - \text{offset}), 1 + \text{offset}]$. Hue is selected from $[-\text{value}, \text{value}]$. The offset and value parameters used are given in Table \ref{augmentation-table-1}
  \item Grayscale: converting the RGB image to a grayscale image with three channels using $(0.2989 \times  r + 0.587 \times g + 0.114 \times  b)$.
  \item Gaussian blurring: smoothing the input image by filtering with a Gaussian kernel. The radius parameter for the kernel is selected uniformly between $0.1$ and $2.0$ pixels.
  \item Solarization: inverting image pixel values by subtracting the maximum value. We keep the result if it is above the threshold value; otherwise, replace it with the original value. The default threshold value is $128$.
\end{itemize}

The spatial dimensions of images input to the encoder networks are $32\times32$ for CIFAR datasets, $64\times64$ for Tiny ImageNet, and $224\times224$ for ImageNet-100 and ImageNet-1K. All pretraining augmentation parameters are listed in Table \ref{augmentation-table-1}.

%\checkmark
% Options: value -> offset, naming in Pytorch 
\begin{table}[h]
  \caption{Augmentation parameters are used in pretraining. Aug-1 and Aug-2 refer to augmentations for each branch. Transformations are selected independently for the two branches with the given probability values. The offset and maximum values determine the interval for uniform selection. } 
  
  \label{augmentation-table-1}
  \centering
  \begin{tabular}{lcccc}
    \toprule
    {Transformation} & \multicolumn{1}{c}{Aug-1} &  \multicolumn{1}{c}{Aug-2}   \\
    %\cmidrule(l){1-1} \cmidrule(l){2-3} \cmidrule(l){4-5} 
    \midrule 
    Random resized cropping probability & 1.0 & 1.0  \\
    Horizontal flipping probability & 0.5 & 0.5  \\
    Color Jitter (CJ) probability & 0.8 & 0.8  \\
    CJ - Brightness offset  & 0.4 & 0.4  \\
    CJ - Contrast offset  & 0.4 & 0.4  \\
    CJ - Saturation offset & 0.2 & 0.2  \\
    CJ - Hue maximum value & 0.1 & 0.1  \\
    Grayscale probabillity & 0.2 & 0.2  \\
    Gaussian blur probability  & 1.0 & 0.1 \\
    Solarization probability   & 0.0 & 0.2\\
    \bottomrule
  \end{tabular}
\end{table}

\subsection{Augmentations during linear evaluation}

 In the linear evaluation stage, a single transformed version of the input image is generated during both the training and the test phases. 
 \begin{itemize}
 \item In the {\it training phase}, the random-resized-crop and horizontal-flip operations are used as augmentations as in \citep{grill2020bootstrap, zbontar2021barlow, bardes2021vicreg, chen2021exploring}. For the random-resized-crop operation, the target sizes are $32\times32$ for CIFAR datasets, $64\times64$ for Tiny ImageNet, and $224\times224$ for ImageNet-100 and ImageNet-1K.
 
\item In the {\it test phase} of ImageNet-100 and ImageNet-1K,  we apply the same preprocessing operation used for the (linear evaluation) test phase of the ImageNet dataset in \citep{grill2020bootstrap}: input images are resized to $256\times 256$ then center cropped to $224\times 224$. For the other datasets, we apply a similar preprocessing by preserving the resize-crop ratio, as in \citep{haochen2021provable}.
\end{itemize}

\subsection{Additional information about augmentation}

As the last step of the augmentation process, we normalize each channel of the resulting tensors by the mean and standard deviation of that channel calculated over the whole input dataset. The mean and standard deviation normalization values for the channels are  $(0.4914, 0.4822, 0.4465)$ and  $(0.247, 0.243, 0.261)$ respectively, for CIFAR-$10$. For CIFAR-$100$, the corresponding normalization values are $(0.5071, 0.4865, 0.4409)$ and $(0.2673, 0.2564, 0.2762)$; for Tiny ImageNet, ImageNet-100 and ImageNet-1K: $(0.485, 0.456, 0.406)$ and $(0.229, 0.224, 0.225)$. 
%Values in parentheses show mean and standard deviation for each channel, respectively. 
As an interpolation method for resizing, we use bicubic interpolation.

\section{Hyper-parameters}
\label{app:parameters}

To select the hyper-parameters, we tested the following values: For CIFAR datasets, we examined $\alpha=[250, 500, 1000]$ , projector output dimensions [$64$, $128$, $256$],  projector hidden dimensions [$2048$, $4096$] with [$2$, $3$] layers setting. For the Tiny ImageNet dataset, we tested $\alpha=[250, 500, 1000]$, the projector dimensions [$4096-4096-128$, $4096-4096-256$], and $\text{learning rate}=[0.25, 0.5, 1.0]$ and the forgetting factor [$0.1$,$0.01$]. For ImageNet-$100$, we found the current best parameters in the same parameters set after the Tiny ImageNet experiments. %\dy{also include ImageNet-1K parameter ranges, fix statement about ImageNet-100}

For ImageNet-1K, we tried $\alpha=[1000, 2000, 3000]$. Using the results of previous experiments in smaller datasets, increasing the output dimension with batch size provides an increase in the test accuracy performance in a limited number of experiments. We get our reported result with $8192-8192-512$ with a batch size of $1536$.

For the covariance update expression in \ref{eq:autocovupdate}, we use the initialization $\hat{\bR}_{\bz^{(q)}}[0]=\mathbf{I}$ for $q=1,2$. We initialize the cross-covariance matrix with $\hat{\bR}_{\bz^{(1)}\bz^{(2)}}[0]=\mathbf{0}$.  We use $\bmu^{(1)}[0]=\bmu^{(1)}[0]= \mathbf{0}$  for the mean initializations. The forgetting factor is $\lambda=0.01$.
Diagonal perturbation constant for the covariance matrices is   $\varepsilon=1\mathrm{e}{-8}$.

Regarding all linear evaluations, we train the linear classifier for $100$ epochs with a batch size of $256$. We use the SGD optimizer with a momentum of $0.9$, no weight decay. For all datasets except ImageNet-$1$K, a learning rate of $0.2$ is chosen. We use the cosine decay learning rate rule as a scheduler with a minimum learning rate of $0.002$. For Imagenet-$1$K, we use a step  scheduler where the learning rate starts with $25$ and reduces by a factor of $10$ for each $20$ epochs.

Below, we summarize the experimental details for each dataset:

\subsection{CIFAR-10 experiment}

For CIFAR-$10$, the encoder network is ResNet-18 modified for CIFAR based on the small  ($32\times32$) input image size. The modifications are: using a smaller kernel size, $3\times 3$ instead of $7\times 7$, in the first convolutional layer, and dropping the max-pooling layer. For the projection block, we use a $3$-layer MLP network with dimensions $2048-2048-64$.

We pretrain for $1000$ epochs using a batch size of $512$ using an SGD optimizer with a momentum of $0.9$, and maximum learning rate of $0.5$. For the learning rate scheduler, we utilize the cosine decay schedule after a linear warm-up for $10$ epochs. During the linear warm-up period, the learning rate starts at $0.003$. At the end of the training, it reaches its minimum value, which is set as $1\mathrm{e}{-6}$. The weight decay parameter for the optimizer is $0.0001$.

\subsection{CIFAR-100 experiment}

For CIFAR-$100$, the encoder network is modified ResNet-18, as explained in the above part. For the projection network, we use a $3$-layer MLP with sizes of $4096-4096-128$.

We pretrain for $1000$ epochs with a batch size of $512$. We use SGD optimizer with a momentum of $0.9$, and a maximum learning rate of $0.5$. As a scheduler, we utilize cosine decay learning rate with linear warm-up for $10$ epochs. The learning rate start value is $0.003$ for the warm-up period. After reaching $0.5$, the learning rate decays to $1\mathrm{e}{-6}$ until the end of the training. The weight decay parameter for the optimizer is $0.0001$.

\subsection{Tiny ImageNet experiment}

For Tiny ImageNet, the encoder network is standard ResNet-$50$. For the projection network, we use a $3$-layer MLP with sizes of $4096-4096-128$.

We pretrain our model for $800$ epochs to make it more comparable with other methods in Table \ref{evaluation-table-1}, and use a batch size of $1024$. We use the SGD optimizer with a momentum of $0.9$, and a maximum learning rate of $0.5$. As a scheduler, we use the cosine decay learning rate with linear warm-up for $10$ epochs. The learning rate start value is $0.003$ for the warm-up period. After reaching $0.5$, the learning rate decays to $1\mathrm{e}{-6}$ until end of the training. The weight decay parameter for the optimizer is $0.0001$.

\subsection{ImageNet-100 experiments}

\subsubsection{ResNet-18}

In this experiment for ImageNet-$100$, the encoder network is standard ResNet-$18$. We use a $3$-layer MLP with sizes of $4096-4096-128$ as the projection network.

We pretrain $400$ epochs to make it more comparable with other methods in Table \ref{evaluation-table-1}, using a batch size of $1024$. We use the SGD optimizer with momentum of $0.9$, and maximum learning rate of $1.0$. As a scheduler, we use the cosine decay learning rate with linear warmup for $10$ epochs. The learning rate start value is $0.003$ for the warm-up period. After reaching $1.0$, the learning rate decays to $0.005$ until the end of the training. The weight decay parameter for the optimizer is $0.0001$.

\subsubsection{ResNet-50}

In this experiment for ImageNet-$100$, the encoder network is standard ResNet-$50$. We use a $3$-layer MLP with sizes of $4096-4096-128$ as the projection network.

We pretrain $200$ epochs to make it more comparable with other methods in Table \ref{evaluation-table-1}, using a batch size of $1024$. We use the SGD optimizer with a momentum of $0.9$, and a maximum learning rate of $1.0$. As a scheduler, we use the cosine decay learning rate with linear warmup for $10$ epochs. The learning rate start value is $0.003$ for the warm-up period. After reaching $1.0$, the learning rate is decayed to $0.005$ until end of training. The weight decay parameter for the optimizer is $0.0001$.

\subsection{ImageNet-1K experiments} \label{app:imagenet1k-hyp}

The encoder network is standard ResNet-$50$ in ImageNet-$1$K experiments. As a projection network in pretraining, a $3$-layer MLP with sizes of $8192-8192-512$ with batch-normalization is utilized. Note that the increase in the number of classes, relative to the Imagenet-100 dataset, translates into an increase in the projector dimension. This further translates into an increased batch size for more accurate estimation of projector covariance matrix with larger dimensions. We pretrain $100$ epochs with a batch size of $1536$. We use the SGD optimizer with a momentum of $0.9$, and a maximum learning rate of $0.2$. As a scheduler, we use the cosine decay learning rate with linear warmup for $10$ epochs. The learning rate start value is $0.003$ for the warm-up period. After reaching $0.2$, the learning rate is decayed to $1\mathrm{e}{-6}$ until end of training. The weight decay parameter for the optimizer is $0.0001$.

\subsubsection{Semi-supervised learning}
In semi-supervised learning, we fine-tune our pretrained model on ImageNet-1K. In contrast to linear evaluation, weights of the encoder also change. Fine-tuning runs $20$ epochs with a batch size of $256$.
For fine-tuning with $1\%$ samples of ImageNet-$1$K, learning rate for the encoder network is $0.005$, and the learning rate for the linear classifier head is $30$. For fine-tuning with $10\%$ samples of ImageNet-$1$K, learning rate for the encoder network is $0.005$, and the learning rate for the linear classifier head is $20$. We used separate step-type learning schedulers for the header and the backbone components, where the learning rate is reduced by a factor of $10$ for the header and a factor of $5$ for the backbone network for every $5$-epoch interval.

We used a learning rate of $0.3$ with a batch size of $2048$ for the VICReg pretraining\citep{bardes2021vicreg}, then used published parameters and VICReg codes\citep{bardes2021vicreg} to evaluate the fine-tuning results.
%All hyperparameters used are reported in \citep{bardes2021vicreg} with the original code. 

\subsection{Hyper-parameter sensitivities}
\label{app:sensitivity}

We provide Top-1 test accuracy results after 1000 epochs of pretraining on the CIFAR-100 dataset for various hyper-parameter adjustments below. The results show that our approach is fairly robust to small hyper-parameter changes. 

\subsubsection{Attraction coefficient} %($\alpha$)

\hspace*{1em}

\begin{table}[h]
  \caption{Test accuracy results for the different attraction coefficients ($\alpha$) with the setup which provides best result for CIFAR-100 dataset.}  
  \label{alpha-table-1}
  \centering
  \begin{tabular}{lccccc}
    \toprule
    Attraction coefficient ($\alpha$) &  {$125$} & {$250$} &  {$500$} & {$1000$} &  {$2000$} \\
    \midrule 
    {Top-1 accuracy} & 69.04 & 71.19 & 71.34 & 71.61 & 69.96\\
    \bottomrule
  \end{tabular}
\end{table}

\subsubsection{Batch size}

\hspace*{1em}

\begin{table}[h]
  \caption{Test accuracy results for the different batch sizes with the setup which provides best result for CIFAR-100 dataset.}
  \label{batchsize-table-1}
  \centering
  \begin{tabular}{lccc}
    \toprule
    Batch sizes  & {$256$} &  {$512$} & {$1024$}  \\
    \midrule 
    {Top-1 accuracy} & 70.32 & 71.61 & 69.98 \\
    \bottomrule
  \end{tabular}
\end{table}

\subsubsection{Learning rate}
\hspace*{1em}

\begin{table}[h]
  \caption{Test accuracy results for the different learning rates with the setup which provides best result for CIFAR-100 dataset.}  
  \label{learningrate-table-1}
  \centering
  \begin{tabular}{lccc}
    \toprule
    Learning rates  & {$0.25$} &  {$0.5$} & {$1.0$}  \\
    \midrule 
    {Top-1 accuracy} &  71.32 & 71.61 & 69.95 \\
    \bottomrule
  \end{tabular}
\end{table}

\subsubsection{Projector Output Dimension}

\begin{table}[h!]
  \caption{Test accuracy results for the different projector output dimensions with the setup which provides best result for CIFAR-100 dataset.}  
  \label{projectordim-table-1}
 \centering
  \begin{tabular}{lccc}
    \toprule
    Projector output dimension  & {$64$} &  {$128$} & {$256$}  \\
    \midrule 
    {Top-1 accuracy} &  68.19 & 71.61 & 71.26  \\
    \bottomrule
  \end{tabular}
\end{table}

\newpage

\section{Algorithm runtime results}
\label{app:runtime}

In Section \ref{sec:complexity} we described the computational complexity of the CorInfoMax approach and stated that its overall effect on runtime is not significant. In this section we provide some experimental evidence.

% Due to using output representations based feature dimension for calculating loss, 
We selected VicReg \citep{bardes2021vicreg} for comparison. We integrated their loss function to our code to eliminate differences in implementation of methods. We ran 10 epochs with both loss functions for different projector output dimensions changing between $64$ and $1024$, and the results are reported in Table \ref{complexity-table-1}.

% 1024 - may check again in a bigger memory GPU
\begin{table}[h]
  \caption{Runtime results for the CIFAR-10 dataset with a batch size of $512$ on a T4 Cloud GPU. Average seconds per epoch from 10 test runs is reported. The loss function of VicReg \citep{bardes2021vicreg} is integrated to our code for comparison.}
  
  \label{complexity-table-1}
  \centering
  \begin{tabular}{lcccc}
    \toprule
    {Projector Dimensions} &  \multicolumn{1}{c}{VicReg}  & \multicolumn{1}{c}{CorInfoMax}  \\
    %\cmidrule(l){1-1} \cmidrule(l){2-3} \cmidrule(l){4-5} 
    \midrule 
    2048-2048-64 & 87.77  & 87.33 \\
    2048-2048-128 & 87.59 & 87.94 \\
    2048-2048-256 & 88.00 & 88.55 \\
    2048-2048-512  & 88.19 & 90.89 \\
    2048-2048-1024  & 89.35 & 102.1 \\
    \bottomrule
  \end{tabular}
\end{table}

Furthermore, we experiment with different batch and projector sizes in order to measure the proportion of time spent on the calculation of the log-determinant in the loss function as shown in Table~\ref{complexity-table-2}. For these experiments, we use Imagenet-1K and ResNet-50 as the encoder on a V100 Cloud GPU on a machine 5 cores of an Intel Xeon Gold 6248 CPU. We find that the cost of the log-determinant operation is at worst around $5.5\%$ of the computation time for the range of hyperparameters we explore. We thus conclude that the log-determinant calculation does not contribute considerably to the overall training cost.

\begin{table}[h]
  \caption{Proportion of time spent in log-determinant calculation for Imagenet-1K and ResNet-50 with varying batch and projector sizes on a V100 Cloud GPU. Average results from 20 test runs is reported.}
  \label{complexity-table-2}
    % \[
    %   \begin{array}{cc|ccccc}
    %     &\multicolumn{1}{c}{\text{Batch Size}} & \multicolumn{5}{c}{\text{Projector Dimension}} \\
    %     && 64 & 128 & 256 & 512 & 1024 \\
    %     \cline{2-7}
    %     & 256 & 0.40\% & 0.56\% & 1.06\% & 2.53\% & 5.54\% \\
    %     \smash{\rotatebox[origin=c]{90}{\text{}}} & 512 & 0.20\% & 0.28\% & 0.53\% & 1.27\% & 2.83\% 
    %   \end{array}
    % \]
    \[
    \begin{array}{@{}l*{5}{c}@{}}
    \toprule
    \text{Batch Size} & \multicolumn{5}{c@{}}{\text{Projector Dimension}}\\
        \cmidrule(l){2-6}
        & 64 & 128 & 256 & 512 & 1024\\
    \midrule
    256 & 0.40\% & 0.56\% & 1.06\% & 2.53\% & 5.54\% \\
    512 & 0.20\% & 0.28\% & 0.53\% & 1.27\% & 2.83\% \\
    \bottomrule
    \end{array}
    \]
\end{table}

\newpage
\section{Full comparison with solo-learn}
\label{app:fullcomparison}

Some of the results in Table \ref{evaluation-table-1} are from the solo-learn paper \citep{da2022solo}. Table \ref{extended-evaluation-table-1} shows the full results from \citep{da2022solo} in comparison to CorInfoMax.

\begin{table}[h!]
  \caption{Top-$1$ and Top-$5$ accuracies {($\%$)} under linear evaluation on CIFAR-$10$, CIFAR-$100$, and ImageNet-$100$ datasets with ResNet-$18$. We bold all top results  that are statistically indistinguishable.  } 
  \label{extended-evaluation-table-1}
  \centering
  \begin{tabular}{lcccccc}
    \toprule
    \multirow{2}[2]{*}{Method} & \multicolumn{2}{c}{CIFAR-10} &  \multicolumn{2}{c}{CIFAR-100} &  \multicolumn{2}{c}{ImageNet-100}  \\
    \cmidrule(l){2-3} \cmidrule(l){4-5} \cmidrule(l){6-7} 
    & {Top-1} & {Top-5} &  {Top-1} & {Top-5}  &  {Top-1} & {Top-5} \\
    \midrule 
    Barlow Twins & 92.10 & 99.73 & 70.90 & \textbf{91.91} & \textbf{80.38} & \textbf{95.28} \\
    BYOL & 92.58 & 99.79 & 70.46 & \textbf{91.96} & \textbf{80.32} & \textbf{94.94} \\
    DeepCluster V2 & 88.85 & 99.58  & 63.61  & 88.09 & 75.40 & 93.22 \\
    DINO & 89.52 & 99.71  &  66.76 & 90.34 & 74.92 & 92.92 \\
    MoCo V2+ & \textbf{92.94} & 99.79 &  69.89 & 91.65 & 79.28 & \textbf{95.50} \\
    NNCLR & 91.88 &99.78  &  69.62 &  91.52 & \textbf{80.16} & \textbf{95.28} \\
    ReSSL & 90.63 &99.62 & 65.92 & 89.73 & 78.48 & 94.24 \\
    SimCLR & 90.74 &99.75 & 65.78 & 89.04 & 77.48 & 94.02 \\
    Simsiam & 90.51 &99.72 & 66.04 & 89.62 & 78.72 & 94.78 \\
    SwAV & 89.17 &99.68 & 64.88 & 88.78 & 74.28 & 92.84 \\
    VICReg & 92.07& 99.74 & 68.54 & 90.83 & 79.40 & \textbf{95.06} \\
    W-MSE & 88.67 &99.68 & 61.33 & 87.26 & 69.06 & 91.22 \\
    \midrule 
    CorInfoMax & \textbf{93.18} & 99.88 &  \textbf{71.61}  &  \textbf{92.40} &  \textbf{80.48} & \textbf{95.46}\\
    \bottomrule
  \end{tabular}
\end{table}

\section{Transfer learning for object detection and instance segmentation example}
\label{app:transfer}

In this section, we outline our experiments on object detection and instance segmentation in order to explore the capability of models trained using our approach to capture fine-grained details, which are otherwise not explicitly explored in classification tasks. We fine-tune our model after 100 epochs of pretraining on ImageNet-1k using Mask R-CNN \citep{he2017maskrcnn} (C4 Backbone) on COCO \citep{lin2014coco}, following the same procedure as MoCo \citep{he2019moco}. To have a reference point in the same exact environment, we finetune the pretrained MoCo V2 \citep{chen2020mocov2} checkpoint for 200 epochs on ImageNet-1K. Table \ref{transfer-evaluation-table} shows the results for both tasks. We achieve similar performance as MoCo V2. We leave further optimization, experimentation on other tasks and datasets, and further exploration of the fine-grained features learned by our model to future work.

\begin{table}[h]
  \caption{AP, AP50, and AP75 for object detection and instance segmentation on COCO. All models have been trained on the 2017 training split and evaluated on the 2017 validation split.} 
  \label{transfer-evaluation-table}
  \centering
  \begin{tabular}{lcccccc}
    \toprule
    \multirow{2}[2]{*}{Method} & \multicolumn{3}{c}{COCO Detection} &  \multicolumn{3}{c}{COCO Segmentation} \\
    \cmidrule(l){2-4} \cmidrule(l){5-7} 
    & {AP\textsubscript{50}} & {AP} &  {AP\textsubscript{75}} & {AP\textsubscript{50}}  &  {AP} & {AP\textsubscript{75}} \\
    \midrule 
    MoCo V2 & 60.52 & 40.77 &  44.19 & 57.33 & 35.56 & 38.12 \\
    % \midrule 
    CorInfoMax & 60.56 & 40.50 &  43.89  &  57.22 &  35.34 & 37.70 \\
    \bottomrule
  \end{tabular}
\end{table}

\newpage
\section{Embedding visualization after pretraining}
\label{app:embeding}
\begin{figure}[h!]
\centering
\includegraphics[width=12.cm, clip]{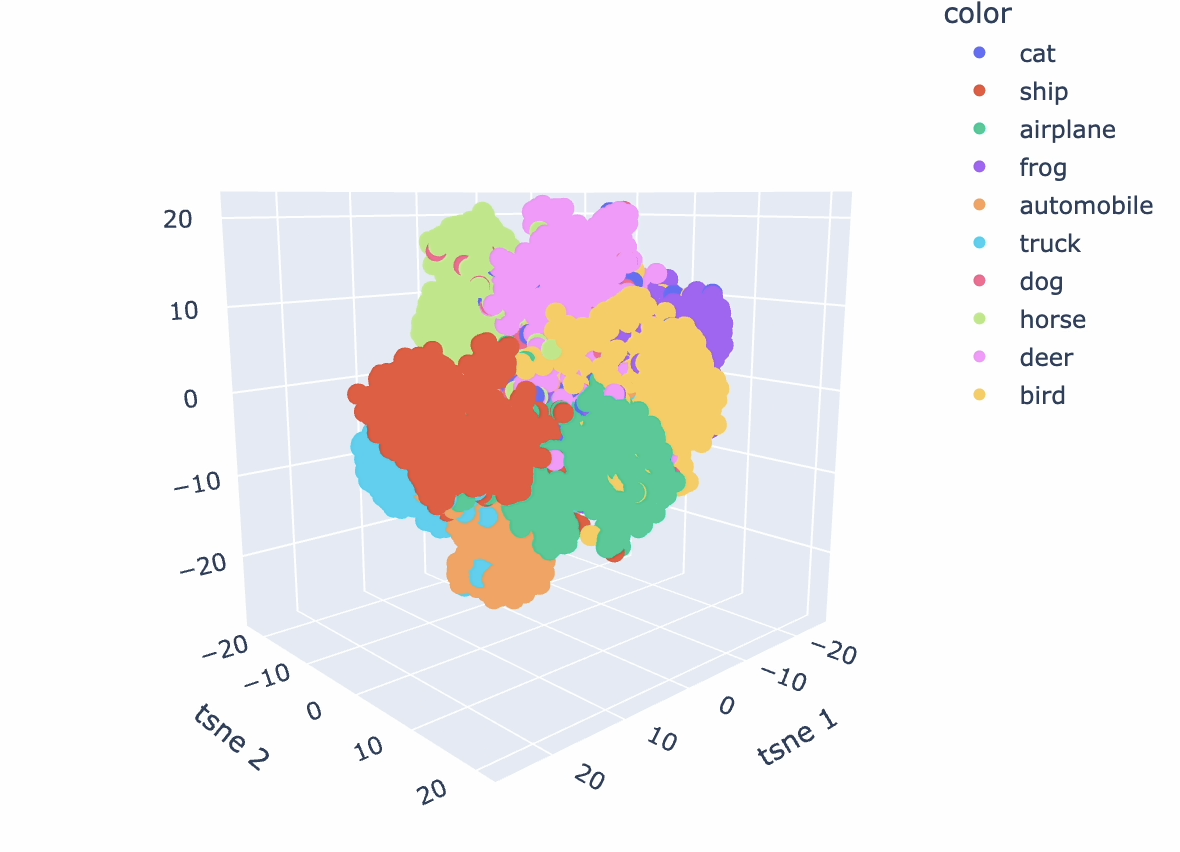}
\caption{t-SNE visualization of obtained embeddings of CIFAR-$10$ test dataset from the output of the encoder network after 1000-epoch pretraining. Each color represents one class of CIFAR-$10$. }
\label{fig:embeddings}
\end{figure}

In Figure \ref{fig:embeddings}, we use t-Distributed Stochastic Neighbor Embedding (t-SNE) to visualize where embeddings of the test dataset are located  after pretraining. We get embeddings from the output of the encoder network using our pretrained model, save them with their classification labels. Then using t-SNE, we visualize the 3D embedded space. We used the following t-sne parameters: perplexity is $50$, early exaggeration is $12$, and iteration number is $1000$, random initialization of embeddings is used, learning rate is $208.33$. We also include a video in the supplementary material that shows the t-SNE plot with changing view angles.

\newpage
\section{Visualization of projector covariance matrix eigenvalues}
\label{app:eigs}
\begin{figure}[h!]
\centering
\includegraphics[width=11.cm, clip]{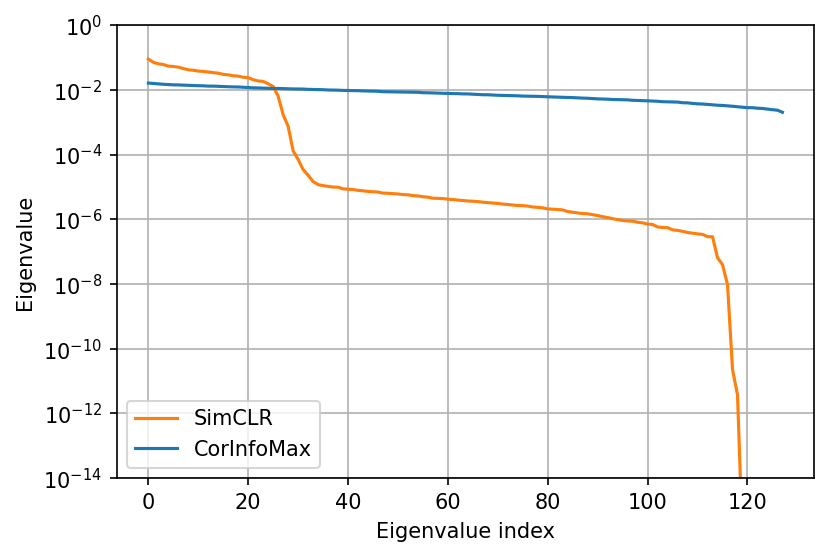}
\caption{Comparison of the sorted eigenvalues of the projector vector covariance matrix $\hat{\bR}_{\bz^{(1)}}$ for CorInfoMax and SimCLR algorithms, for CIFAR-10 dataset and projector dimension of 128.}
\label{fig:eigplot}
\end{figure}

The eigenvalues of the projector vector covariance matrix, $\hat{\bR}_{\bz^{(1)}}$ reflect the effective use of the embedding space. Due to the existence of the "big-bang factor", $\log\det(\hat{\bR}_{\bz^{(1)}}[l]+\varepsilon\mathbf{I})$, in (\ref{eq:covdet2}), we expect that no dimensional collapse occurs (with very small $\varepsilon$), and eigenvalues take significant values.
To confirm this expectation, we performed the following experiment: for the CIFAR-10 dataset, we ran the contrastive SimCLR algorithm and obtained its projector covariance matrix after training. Similarly, we naturally obtained the projector covariance matrix for the CorInfoMax approach. For both algorithms, we used a projector dimension of $128$. Figure \ref{fig:eigplot} compares the sorted eigenvalues of both covariances. As can be observed from this figure, the effective embedding space dimension for the SimCLR algorithm is small, most of the energy is concentrated at the first $50$ eigenvalues. Furthermore, the smallest $8$-eigenvalues are equal to $0$, within numerical precision, indicating a dimensional collapse. On the contrary, the eigenvalues for the CorInfoMax algorithm are significant for all dimensions, hinting at the effective use of the embedding space. 

\newpage
\section{Visualization of LD-mutual information evolution during pretraining}
\label{app:minfoevolution}
\begin{figure}[h!]
\centering
\includegraphics[width=11.cm, clip]{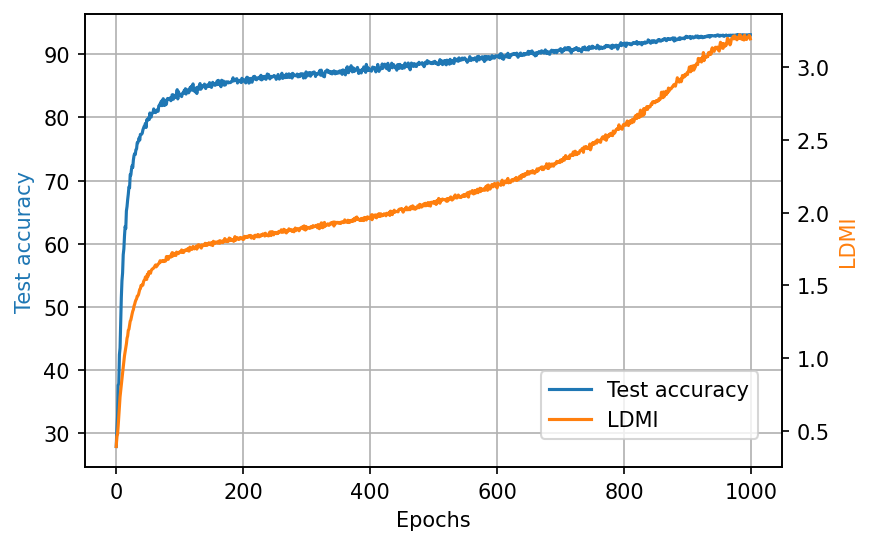}
\caption{Evolution of the LDMI measure and the test accuracy for the CIFAR-10 dataset as a function of the CorInfoMax algorithm epochs.}
\label{fig:cifar10ldmiplot}
\end{figure}

As the CorInfoMax objective function is derived from the LDMI measure, it would be interesting to observe its evolution in relation to algorithm epochs and (online) test accuracy. For this purpose, we performed two experiments with the CIFAR-10 and CIFAR-100 datasets, where we used the CorInfoMax loss function, but recorded the Training-LDMI values (based on the covariance estimates using (\ref{eq:autocovupdate}) during training) and the test accuracy values. Figure \ref{fig:cifar10ldmiplot} shows the progress of the Training-LDMI and the test accuracy curves on the same graph for the CIFAR-10 dataset.

\begin{figure}[h!]
\centering
\includegraphics[width=11.cm, clip]{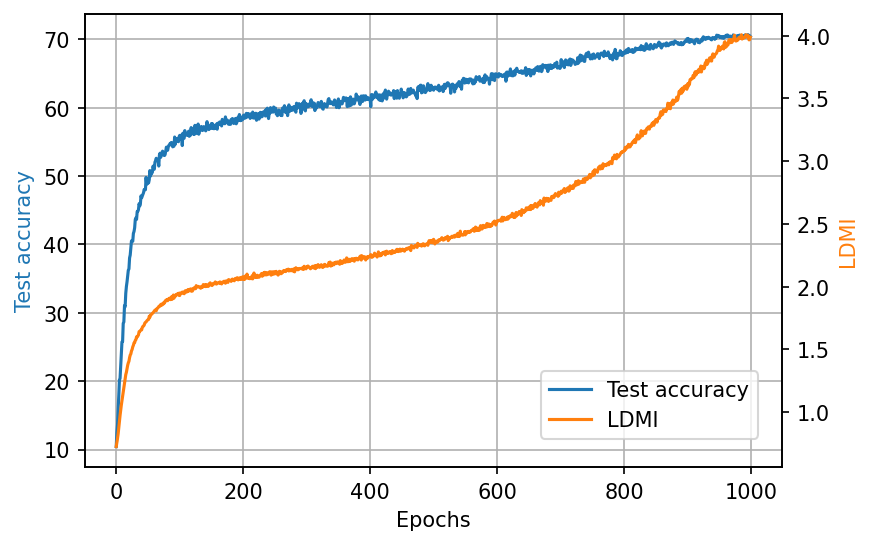}
\caption{Evolution of the LDMI measure and the  test accuracy for the CIFAR-100 dataset as a function of the CorInfoMax algorithm epochs.}
\label{fig:cifar100ldmiplot}
\end{figure}

Similarly, Figure \ref{fig:cifar100ldmiplot} shows the progress of the Training-LDMI and the test accuracy curves on the same graph for the CIFAR-100 dataset.

Both figures confirm that the Training-LDMI measure and the test accuracy increase together almost monotonically, ignoring the small variations.  

\section{Supplementary notes on CorInfoMax criterion}
\label{app:suppcorinfomax}
\subsection{Gradients of the CorInfoMax objective}
Let $\bz^{(q)}_n$  represent the $n^{th}$ sample of the $q^{th}$ branch projector output for the current batch, where $n \in \{1, \ldots, N\}$ and $q=1,2$. We provide the gradient expressions of the CorInfoMax objective (\ref{eq:covdet2}) with respect to the projector outputs, which are backpropagated to the train the encoder networks.

\begin{figure}[t]
\centering
\includegraphics[width=10cm, trim=17.5cm 10.0cm 1.5cm 1.0cm,clip]{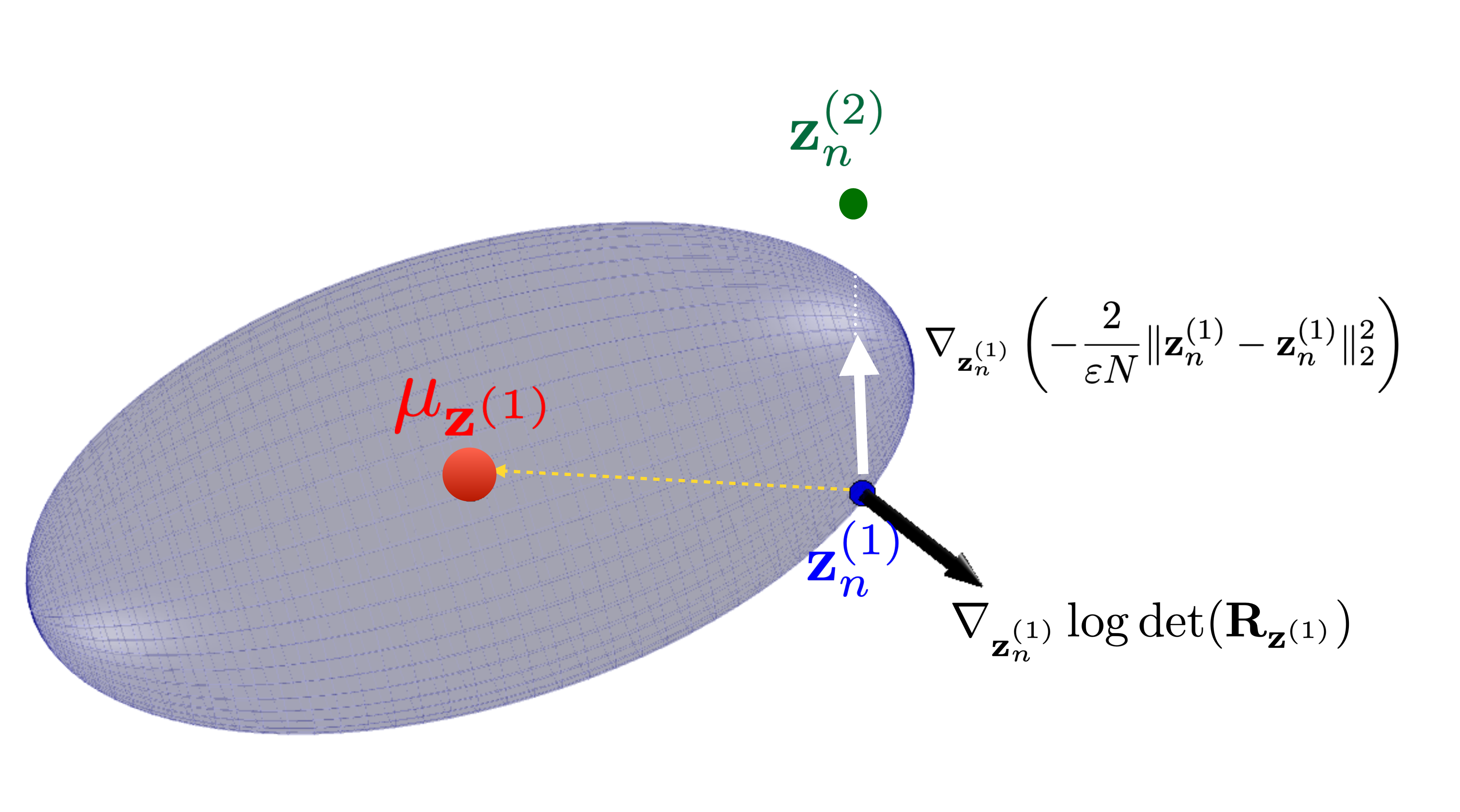}
\caption{Gradients of the objective function for CorInfoMax:  The ellipsoid is the level surface of the quadratic function $q(\bz)$ in (\ref{eq:levelellispsoid}) containing $\bz^{(1)}_n$, one of the projector-$1$ output samples, the black arrow represents the gradient of  $\log\det(\bR_{\bz^{(1)}})$ with respect to $\bz^{(1)}_n$, the white arrow represents the gradient of $\frac{2}{\varepsilon N}\|\bz_n^{(1)}-\bz_n^{(2)}\|_2^2$ with respect to $\bz_n^{(1)}$. } 
\label{fig:CImaxGrads}
\end{figure}

For the first term on the right side of (\ref{eq:covdet2}), we can write
\begin{eqnarray}
&&\nabla_{\bz^{(1)}_n}\log\det(\hat{\bR}_{\bz^{(1)}}[l]+\varepsilon \mathbf{I})=\frac{(1-\lambda)}{N}(\hat{\bR}_{\bz^{(1)}}[l]+\varepsilon\mathbf{I})^{-1}(\bz^{(1)}_n-\bmu_{\bz^{(1)}}[l]), \label{eq:G11}
\end{eqnarray}
and, 
\begin{eqnarray*}
&&\nabla_{\bz^{(2)}_n}\log\det(\hat{\bR}_{\bz^{(1)}}[l]+\varepsilon \mathbf{I})=0.
\end{eqnarray*}

Similarly, for the second term on the right side of (\ref{eq:covdet2}), we can write
\begin{eqnarray*}
&&\nabla_{\bz^{(2)}_n}\log\det(\hat{\bR}_{\bz^{(2)}}[l]+\varepsilon \mathbf{I})=\frac{(1-\lambda)}{N}(\hat{\bR}_{\bz^{(2)}}[l]+\varepsilon\mathbf{I})^{-1}(\bz^{(2)}_n-\bmu_{\bz^{(2)}}[l]), \hspace{0.1in}
\end{eqnarray*}
and, 
\begin{eqnarray*}
&&\nabla_{\bz^{(1)}_n}\log\det(\hat{\bR}_{\bz^{(2)}}[l]+\varepsilon \mathbf{I})=0.
\end{eqnarray*}
Finally, for the rightmost term of (\ref{eq:covdet2}), which is the Euclidian distance based loss, we can write
\begin{eqnarray}
\nabla_{\bz^{(1)}_n}\left(-\frac{2}{\varepsilon N}\|\bZ^{(1)}[l]-\bZ^{(2)}[l]\|_F^2\right)=\frac{4}{\varepsilon N}(\bz_n^{(2)}-\bz_n^{(1)}), \label{eq:G21}
\end{eqnarray}
and
\begin{eqnarray}
\nabla_{\bz^{(2)}_n}\left(-\frac{2}{\varepsilon N}\|\bZ^{(1)}[l]-\bZ^{(2)}[l]\|_F^2\right)=\frac{4}{\varepsilon N}(\bz_n^{(1)}-\bz_n^{(2)}). \label{eq:G22}
\end{eqnarray}

Inspecting the gradient expressions with respect to $\bz^{(1)}_n$: The vector in (\ref{eq:G11}) is the surface normal of the level set of the quadratic function 
\begin{eqnarray}
q(\bz)=\frac{(1-\lambda)}{N}(\bz-\bmu_{\bz^{(1)}}[l])^T(\hat{\bR}_{\bz^{(1)}}[l]+\varepsilon\mathbf{I})^{-1}(\bz-\bmu_{\bz^{(1)}}[l]),  \label{eq:levelellispsoid}
\end{eqnarray}
at $\bz=\bz^{(1)}_n$, which is illustrated by the black vector in Figure \ref{fig:CImaxGrads}. Since $(\hat{\bR}_{\bz^{(1)}}[l]+\varepsilon\mathbf{I})^{-1}$ is positive definite, the inner product
\begin{eqnarray}
&&\langle \nabla_{\bz^{(1)}_n}\log\det(\hat{\bR}_{\bz^{(1)}}[l]+\varepsilon \mathbf{I}), \bmu_{\bz^{(1)}}[l]-\bz^{(1)}_n\rangle \nonumber \\
&&=- \frac{(1-\lambda)}{N}(\bz^{(1)}_n-\bmu_{\bz^{(1)}}[l])^T(\hat{\bR}_{\bz^{(1)}}[l]+\varepsilon\mathbf{I})^{-1}(\bz^{(1)}_n-\bmu_{\bz^{(1)}}[l])\nonumber%\\
%&<& 0,
\end{eqnarray}
is negative,  which implies that the vector pointing from $\bz_n^{(1)}$ towards the center $\mu_{z^{(1)}}[l]$, the yellow dashed arrow in Figure \ref{fig:CImaxGrads}, makes an obtuse angle with the gradient  $\nabla_{\bz^{(1)}_n}\log\det(\hat{\bR}_{\bz^{(1)}}[l]+\varepsilon \mathbf{I})$.  Therefore, the gradient in (\ref{eq:G11}) is pointing away from the center of the ellipsoid, encouraging the expansion. 

On the other hand, the gradient expressions in (\ref{eq:G21}) and (\ref{eq:G22}) correspond to force  pulling the positive samples $\bz^{(1)}_n$ and $\bz^{(2)}_n$ towards each other as indicated by the white arrow.

A video animation of sample points moving under these gradients is provided in the supplementary material.

%% file: blockdiagram.tex
\begin{center}
%\begin{scaletikzpicturetowidth}{\textwidth}
%\resizebox{\textwidth}{!}{%
{
\begin{tikzpicture}[thick,scale=0.8, every node/.style={transform shape}]
%\begin{tikzpicture}[thick,scale=0.8, every node/.style={scale=0.8}]
  % The order of blocks matters since some are partially hidden behind subsequent blocks.
  \node[inputbnew,text width=1cm](input1){{\large Input \\ \vspace{0.1in}\hspace{0.03in} {\Large $\mathcal{X}$}}};
  \Dimline[($(input1)+(-0.25,2)$)][($(input1)+(1.25,2)$)][{\large N}];
  \DimlineN[($(input1)+(-1.05,1.2)$)][($(input1)+(-1.05,-1.2)$)][{\large M}];
  %\DimlineK[($(input1)+(0.85,-1.3)$)][($(input1)+(1.35,-0.8)$)][{ K}];
  \DimlineKneww[($(input1)+(0.85,-1.3)$)][($(input1)+(1.35,-0.8)$)][{ K}];
  %\node[block,right=1cm of input1,yshift=-2.5cm](augm1){{Augmentation 2}};
  \node[draw,
    fill=coldblue,
    minimum width=2cm,
    minimum height=1.2cm,
    right=1.1cm of input1,
    yshift=75.5,rounded corners](augm1) {\color{black}\LARGE $a_1(\cdot)$};
    \node[right=-2.3cm of augm1, yshift=25]{Augmentation 1};
  \node[draw,
    fill=coldblue,
    minimum width=2cm,
    minimum height=1.2cm,
    right=1.1cm of input1,
    yshift=-65.5,rounded corners](augm2) {\color{black}\LARGE $a_2(\cdot)$};
    \node[right=-2.3cm of augm2, yshift=-25]{Augmentation 2};
    \node[right=0.1cm of input1](l1s){};
    \node[right=0.65cm of input1](l1e){};
    \draw[-,black,thick] (l1s.center) -- (l1e.center);
    \node[right=0.65cm of input1,yshift=75.5](l2e){};
    \node[right=0.1cm of l2e](l3e){};
    \draw[-,black,thick] (l1e.center) -- (l2e.center);
    \draw[->,black,thick] (l2e.center) -- (augm1.west);
    \node[right=0.65cm of input1,yshift=-65.5](l4e){};
    \draw[-,black,thick] (l1e.center) -- (l4e.center);
    \node[right=0.1cm of l4e](l5e){};
    \draw[->,black,thick] (l4e.center) -- (augm2.west);
    \node[inputbAnew,text width=1cm,right=0.0cm of augm2,yshift=65.5](inputa1){{{\Large $\mathcal{X}^{(1)}$}}};
    \node[inputbBnew,text width=1cm,right=1.45cm of augm2,yshift=65.5](inputa2){{{\Large $\mathcal{X}^{(2)}$}}};
    % Augmentation 1 output lines
    \node[right=1.75cm of l3e](l6b){};
    \node[right=2.8cm of l3e](l6e){};
    \draw[-,black,thick] (l6b.center) -- (l6e.center);
    \node[below=0.95cm of l6e](l7e){};
    \draw[->,black,thick] (l6e.center) -- (l7e.center);
    % Augmentation 2 output lines
    \node[right=1.75cm of l5e](l8b){};
    \node[right=3.8cm of l5e](l8e){};
    \draw[-,black,thick] (l8b.center) -- (l8e.center);
    \node[above=0.85cm of l8e](l9e){};
    \draw[->,black,thick] (l8e.center) -- (l9e.center);
    \node[draw,
    fill=coldblue,
    minimum width=2cm,
    minimum height=1.2cm,
    right=1cm of inputa2,
    yshift=2,rounded corners](dnn) {\color{black}\Large $f(\cdot; \mathbb{W}_{DNN})$};
    \node[right=-1.8cm of dnn, yshift=25]{DNN};
    \node[right=0.1cm of inputa2,yshift=2](l10s){};
    \node[right=0.9cm of inputa2,yshift=2](l10e){};
    \draw[->,black,thick] (l10s.center) -- (l10e.center);
    \node[draw,
    fill=coldblue,
    minimum width=2cm,
    minimum height=1.2cm,
    right=0.5cm of dnn,
    yshift=0,rounded corners](proj) {\color{black}\Large $p(\cdot; \mathbb{W}_{P})$};
    \node[right=-1.8cm of proj, yshift=25]{Projector};
    \draw[->,black,thick] (dnn.east) -- (proj.west);
    \node[draw,
    top color=coldgreen2,
    bottom color=coldgreen2,
    minimum width=1.5cm,
    minimum height=2cm,
    right=0.5cm of proj](pout1){\color{black}\large$\mathbf{Z}^{(1)}$};
    \draw[->,black,thick] (proj.east) -- (pout1.west);
    \node[draw,
    top color=coldorange,
    bottom color=coldorange,
    minimum width=1.5cm,
    minimum height=2cm,
    right=0.0cm of pout1](pout2){\color{black}\large$\mathbf{Z}^{(2)}$};
    \DimlineP[($(pout2)+(1.05,1.05)$)][($(pout2)+(1.05,-1.05)$)][{\large P}];
   \Dimline[($(pout1)+(-0.69,1.15)$)][($(pout1)+(0.75,1.15)$)][{\large N}];
    \Dimline[($(pout2)+(-0.69,1.15)$)][($(pout2)+(0.75,1.15)$)][{\large N}];
    % Classifier
  %  \node[right=0.11cm of dnn,yshift=3](c1s){};
   % \node[right=-0.25cm of c1s,yshift=-70](c1e){};
%    \draw[dotted,red,thick] (c1s) -- (c1e);
%    \node[draw,
 %   fill=junglegreen,
  %  minimum width=2cm,
   % minimum height=1.2cm,
%    right=0.3cm of c1e,
 %   yshift=0](class) {\color{white}\Large $c(\cdot; \mathbb{W}_{C})$};
  %  \draw[dotted,->,black,thick] (c1e.center) -- (class.west);
   % \node[right=-2.3cm of class, yshift=-25]{Linear Classifier};
\end{tikzpicture}
}
%\end{scaletikzpicturetowidth}
\end{center}